%% file: neurips_2025.tex
\documentclass{article}
\PassOptionsToPackage{numbers,compress}{natbib}
\usepackage[final]{neurips_2025}

\usepackage[T1]{fontenc}
\usepackage[utf8]{inputenc}
\usepackage{microtype}
\usepackage{fontawesome5}
\usepackage{amsmath,amssymb,amsfonts}
\usepackage{nicefrac}
\usepackage{booktabs}
\usepackage{graphicx}
\usepackage{url}
\usepackage{enumitem}
\usepackage{wrapfig}
\usepackage{float}
\usepackage{multirow}
\usepackage{makecell}
\usepackage{pifont}
\usepackage{csquotes}
\usepackage{adjustbox}
\usepackage{svg}        
\usepackage{xspace}
\usepackage{bm}
\usepackage{tabulary}
\usepackage{pgf,pgfplots}
\pgfplotsset{compat=1.18}
\usepackage{nicematrix}
\usepackage{tikz}
\usepackage[table,dvipsnames]{xcolor}
\usepackage{colortbl}
\usepackage{subcaption}
\usepackage[colorlinks,
    linkcolor=logoBlue,
    citecolor=logoBlue,
    urlcolor=logoBlue]{hyperref}
\usepackage[nameinlink,capitalize,noabbrev]{cleveref}
\usepackage[most]{tcolorbox}

\definecolor{logoBlue}{HTML}{1A4E8A}
\definecolor{logoCyan}{HTML}{56BBCC}
\definecolor{genLLM}{RGB}{230,230,230} 
\definecolor{specLLM}{RGB}{252,228,214}  
\definecolor{ourLLM}{RGB}{230,242,230}  
\definecolor{spaLLM}{RGB}{255,249,196}  
\definecolor{genVLM}{RGB}{255,255,255}   
\definecolor{ourVLM}{RGB}{207,234,242}   
\definecolor{mypurple}{RGB}{184, 36, 255}
\newcommand{\ie}{\textit{i.e.},\xspace}      % i.e.
\newcommand{\eg}{\textit{e.g.},\xspace}      % e.g.
\def\vs{\emph{vs}.\ }                        % versus

\crefname{equation}{Eq.}{Eqs.}
\crefformat{section}{\S#2#1#3}
\crefformat{subsection}{\S#2#1#3}
\crefformat{subsubsection}{\S#2#1#3}
\crefrangeformat{section}{\S\S#3#1#4 to~#5#2#6}
\crefmultiformat{section}{\S\S#2#1#3}{ and~#2#1#3}{, #2#1#3}{ and~#2#1#3}

\tcbuselibrary{breakable, skins}
\newtcolorbox{planbox}[1]{
  enhanced,
  colback=white,
  colframe=logoCyan!80,
  coltitle=logoBlue,
  fonttitle=\bfseries\sffamily,
  title=#1,
  titlerule=0.8pt,
  boxrule=1pt,
  left=1mm, right=1mm, top=1mm, bottom=1mm,
  boxsep=1mm,
  width=1.1\textwidth,
  before upper=\smallskip,
}

\newcommand{\tightcolorbox}[2]{%
  \begingroup
  \setlength{\fboxsep}{1pt}%
  \colorbox{#1}{#2}%
  \endgroup
}

\newcommand{\modelname}{{\textcolor{logoBlue}{Spatial}\textcolor{logoCyan}{Reasoner-R1}}}
\newcommand{\modelnamenc}{{SpatialReasoner-R1}}

\newcommand{\dponame}{{fDPO}}
\newcommand{\mctsname}{M3CTS}
\definecolor{darkgreen}{RGB}{0,130,0}
\definecolor{darkred}{RGB}{180,0,0}

\pgfplotsset{compat=1.18}

\title{%
\vspace{-0.2cm}
  \begin{tabular}{@{} m{1.8cm} m{13cm} @{}}
    \raisebox{-1em}{\includegraphics[width=2cm]{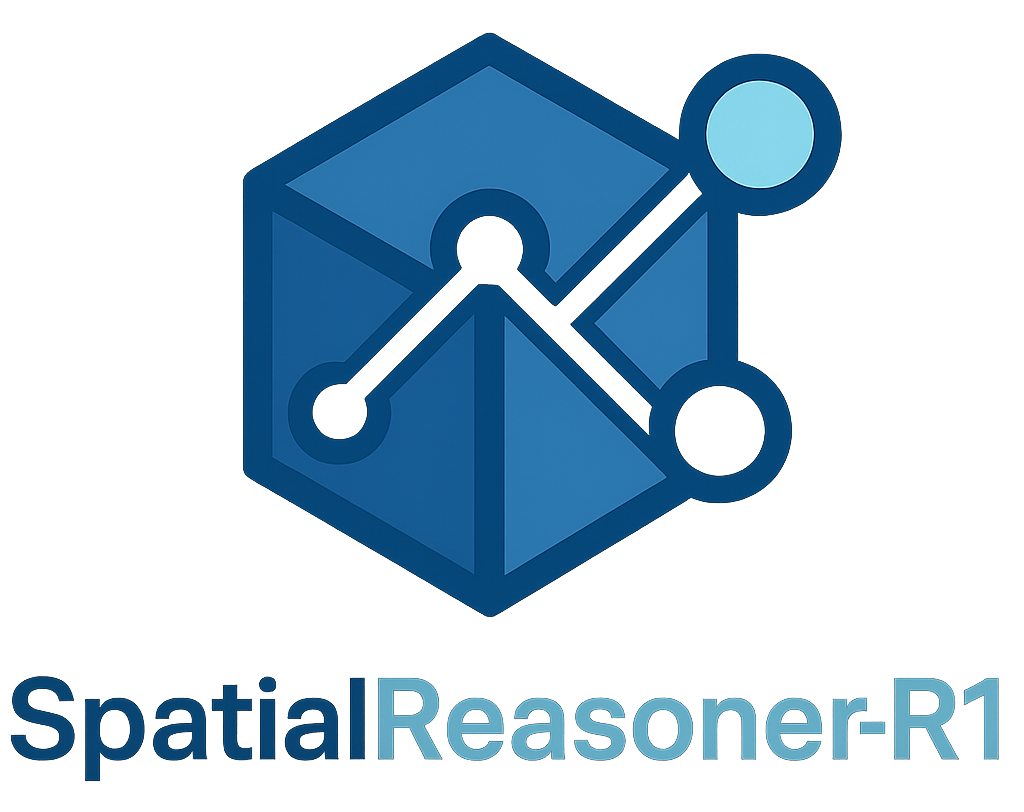}} &
    \begin{tabular}[t]{@{}l@{}}
      \textbf{Fine-Grained Preference Optimization} \\ 
      \textbf{Improves Spatial Reasoning in VLMs}
    \end{tabular}
  \end{tabular}
}

\makeatletter
\renewcommand{\@author}{%
  \textbf{Yifan Shen\textsuperscript{1}, Yuanzhe Liu\textsuperscript{2}, Jingyuan Zhu\textsuperscript{2}, Xu Cao\textsuperscript{1}, Xiaofeng Zhang\textsuperscript{3}, Yixiao He\textsuperscript{1},} \\
  \textbf{Wenming Ye\textsuperscript{4}, James M. Rehg\textsuperscript{1}, Ismini Lourentzou\textsuperscript{1}} \\
  {\textsuperscript{1}University of Illinois Urbana-Champaign 
  \textsuperscript{2}University of Pennsylvania} \\
  {\textsuperscript{3}Shanghai Jiao Tong University
  \textsuperscript{4}Google}\\
  {\texttt{\{yifan26,lourent2\}@illinois.edu}}\\
  \raisebox{-1pt}{\textcolor{mypurple}{\texttt{https://plan-lab.github.io/spatialreasoner}}}\\ 
}
\makeatother

\begin{document}

\maketitle

\vspace{-6mm}
\begin{figure}[ht]
    \centering
    \makebox[\linewidth]{\includegraphics[width=\linewidth]{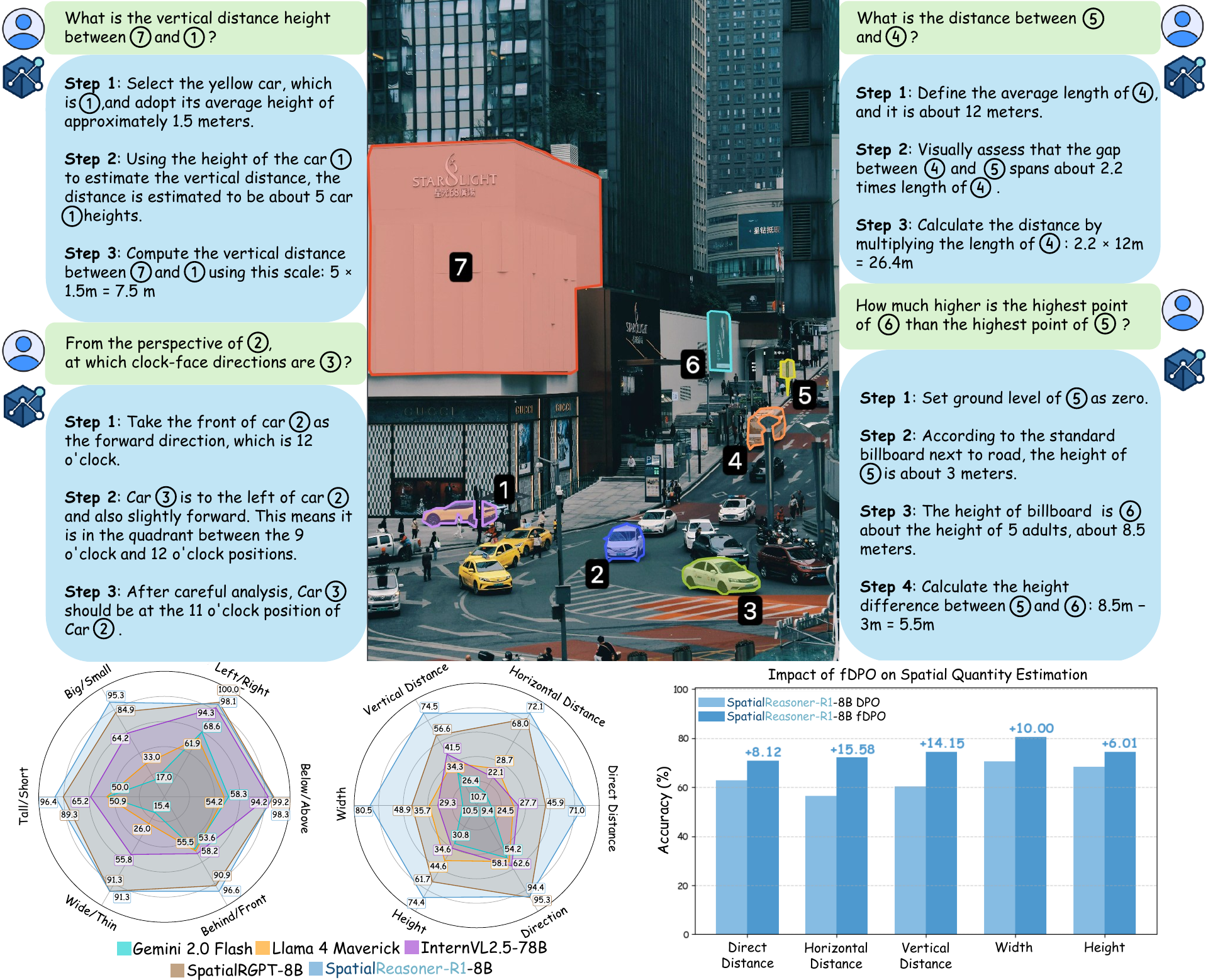}}
    \label{fig:teaser}
\end{figure}
\vspace{-3mm}

\begin{abstract}

Current Vision-Language Models (VLMs) struggle with fine-grained spatial reasoning, particularly when multi-step logic and precise spatial alignment are required. In this work, we introduce \textbf{\modelname{}}, a vision-language reasoning model designed to address these limitations. To construct high-quality supervision for spatial reasoning, we design a Multi-Model Monte Carlo Tree Search (\mctsname{}) method that generates diverse, logically consistent Long Chain-of-Thought (LongCoT) reasoning trajectories. In addition, we propose a fine-grained Direct Preference Optimization (\dponame{}) method that introduces segment-specific preference granularity for descriptive grounding and logical reasoning, guided by a spatial reward mechanism that evaluates candidate responses based on visual consistency, spatial grounding, and logical coherence.
Experimental results demonstrate that \dponame{} achieves relative performance gains of 4.1\% and 9.0\% over standard DPO on spatial qualitative and quantitative tasks, respectively. \modelnamenc{}, trained with \dponame{}, sets a new SoTA on \textsc{SpatialRGPT-Bench}, outperforming the strongest baseline by 9.4\% in average accuracy, while maintaining competitive performance on general vision-language tasks.

\end{abstract}

\input{sections/01_introduction}
\input{sections/02_related_work}
\input{sections/03_method}
\input{sections/04_experiments}
\input{sections/05_conclusion}

\clearpage
\bibliographystyle{plainnat}
\bibliography{neurips_2025}

\clearpage
\input{sections/06_appendix}

%%%%%%%%%%%%%%%%%%%%%%%%%%%%%%%%%%%%%%%%%%%%%%%%%%%%%%%%%%%%

\clearpage

\end{document}

%% file: sections/01_introduction.tex
\section{Introduction}
\label{sec:introduction}

Vision-Language Models (VLMs) have demonstrated significant advancements in multimodal understanding tasks, such as image captioning, visual question answering, object detection, and video interpretation~\cite{antol2015vqa, li2024llava, radford2021learning, team2023gemini, zhang2024mm}. However, their ability to perform spatial reasoning remains limited, especially in scenarios involving complex object arrangements and occlusions~\cite{chen2024spatialvlm, cheng2024spatialrgpt, dong2024insight, yang2024thinking}. This gap poses a significant challenge for applications such as robotics, autonomous driving, and augmented reality, where robust spatial understanding is essential for effective decision-making~\cite{pan2025metaspatial}.

Historically, early VLMs predominantly employed direct-response paradigms~\cite{antol2015vqa, radford2021learning}, \ie producing immediate answers without explicit reasoning, which often leads to shallow understanding. Recent advances in Chain-of-Thought (CoT) prompting have introduced step-by-step reasoning ~\cite{wei2022chain}, but standard CoT traces are often too brief or abstract to capture fine-grained spatial logic. 
In contrast, Long Chain-of-Thought (LongCoT) prompting produces richer, more interpretable reasoning paths that better support comprehension~\cite{chen2025towards, luo2025adar1,  wen2025light}. 
Still, such prompting must go beyond simple depth estimation, as accurate spatial reasoning requires understanding occlusions, relative orientations, and positional ambiguity, all of which are difficult to capture without structured, fine-grained supervision.

To address these challenges, we introduce \textbf{\modelname{}}, a novel VLM designed to perform spatial reasoning directly from 2D images. \modelnamenc{} employs structured, interpretable LongCoT reasoning to systematically parse and solve spatial queries without relying on additional modalities or external sensor data. To optimize the training process for multi-step reasoning, we introduce a new \textbf{fine-grained Direct Preference Optimization (\dponame{})} method that applies differentiated learning updates tailored to two semantically distinct components, descriptive grounding and logical reasoning. Unlike traditional DPO, \dponame{} introduces segment-specific preference granularity, allowing \modelnamenc{} to adjust its optimization for each generation phase, emphasizing spatial localization during descriptive grounding and enhancing multi-step logical inferences during reasoning.\looseness-1 

To curate diverse high-quality spatial reasoning data for training, we propose a \textbf{Multi-Model Monte Carlo Tree Search (\mctsname{})} that generates high-quality LongCoT responses by leveraging collaborative exploration across multiple VLMs, and a \textbf{fine-grained spatial reward mechanism} that evaluates candidate responses across three dimensions: descriptive accuracy, spatial grounding precision, and logical coherence, which are then used to construct positive and negative sample pairs for DPO and fDPO training. Empirical results across several challenging spatial reasoning tasks demonstrate that \modelnamenc{} achieves state-of-the-art performance, significantly outperforming existing VLMs and CoT-based methods, particularly on complex, multi-step spatial reasoning. Specifically, \modelnamenc{} surpasses the best baseline by 9.4\% in average accuracy on spatial understanding. Our \dponame{} improves by 4.1\% and 9.0\% on average over standard DPO on spatial qualitative and quantitative tasks, respectively. 
Our contributions are as follows:
\begin{itemize}[itemsep=0.5ex, parsep=0pt, topsep=0pt]
\item[\textbf{(1)}] We introduce \modelnamenc{}, a LongCoT spatial reasoning VLM that effectively generates interpretable, step-by-step explanations directly from 2D images.  \modelnamenc{} establishes a new SoTA in spatial understanding, while maintaining robust performance on general vision-language benchmarks.

\item[\textbf{(2)}] To enhance training stability and precision, we propose a new fine-grained Direct Preference Optimization (\dponame{}) method that employs segment-specific learning updates tailored explicitly for descriptive grounding and logical reasoning.

\item[\textbf{(3)}] To address the scarcity of high-quality spatial reasoning data, we introduce a data generation pipeline that combines Multi-Model Monte Carlo Tree Search  (\mctsname{}) with fine-grained spatial rewards, enabling the creation of diverse, logically consistent LongCoT trajectories for fine-grained preference training.
\end{itemize}

%% file: sections/02_related_work.tex
\section{Related Work}
\label{sec:related_works}
\textbf{Vision Language Models and Spatial Reasoning.} 
Recent advances in VLMs have significantly enhanced the ability of multimodal models to understand and generate descriptive text grounded in visual contexts~\cite{li2024llava,liu2023visual,liu2024improved, nguyen2025calico, wahed2024prima, zhang2024mm}. Models such as Flamingo~\cite{alayrac2022flamingo}, BLIP-2~\cite{li2023blip}, and Qwen-VL~\cite{liu2023visual} use high-capacity vision encoders~\cite{radford2021learning} paired with LLMs~\cite{brown2020language, touvron2023llama} to achieve state-of-the-art performance in various multimodal tasks, such as visual question answering, image captioning, and instruction following~\cite{antol2015vqa, dai2023instructblip, lin2014microsoft, tu2025longwriter, wu2024multimodal, zhou2024calibrated}. Current trends involve scaling models to improve general understanding~\cite{chen2024internvl, hurst2024gpt, team2023gemini} and using large-scale instruction tuning datasets~\cite{liu2024improved,rasheed2024glamm,yue2024mammoth2}. Both proprietary~\cite{google-deepmind-gemini2024,intelligence2024amazon,hurst2024gpt} and open-source VLMs~\cite{chen2024internvl,deitke2024molmo,yuan2025sa2va} have shown impressive results.

While VLMs show promise in visual understanding, accurately perceiving and reasoning about spatial arrangements remains a challenge~\cite{cheng2024spatialrgpt}.
Recent efforts to improve spatial understanding include fine-tuning VLMs on spatial VQA datasets~\cite{chen2024spatialvlm,chen2023shikra, cheng2024spatialrgpt, liu2025spatialcot, wu2024mind, kong2025autospatial,ranasinghe2024learning}, and zero-shot frameworks that leverage external 3D foundation models for geometric priors~\cite{ma2024spatialpin}. Region-aware models have also been proposed for better grounding and finer spatial queries~\cite{guo2024regiongpt, you2023ferret, yuan2024osprey}. These advances extend to scenarios such as video understanding~\cite{yang2024thinking} and 3D generation~\cite{ma2025spatialreasoner, pan2025metaspatial}. To track progress, specialized benchmarks like Q-Spatial Bench~\cite{liao2024reasoning}, SpatialRGPT-Bench~\cite{cheng2024spatialrgpt}, VSI-Bench~\cite{yang2024thinking}, and 3DSRBench~\cite{ma20243dsrbench} have been introduced to assess spatial skills.
However, current models still struggle with complex, multi-step spatial reasoning. \modelnamenc{} addresses this gap by introducing fine-grained preference optimization and multi-level reward mechanisms.

\textbf{Aligning VLMs using Preference Optimization.} Preference-based learning methods, particularly DPO~\cite{rafailov2023direct}, have become standard techniques for aligning models with human intentions. These methods bypass the need for explicit reward model training and have often demonstrated strong performance compared to earlier Reinforcement Learning with Human Feedback (RLHF) approaches~\cite{azar2024general, ethayarajh2024model, munos2023nash, zhao2023slic}. 
In the multimodal domain, DPO and its variants have been adapted to address specific challenges such as reducing hallucinations and improving visual grounding~\cite{wang2024enhancing, xing2025re, yu2024rlhf}. The adaptability of DPO is further highlighted by its recent application in aligning generative models beyond language, such as text-to-image diffusion models~\cite{gu2024diffusion, li2024aligning, wallace2024diffusion, yang2024using, yuan2024self}. Adaptation methods often involve constructing preference pairs based on human corrections, AI feedback, or contrasting inputs to guide the model towards desired behaviors~\cite{chen2025visrl, cui2024fine, dong2024insight, fu2024mitigating, sun2023aligning, wang2024mdpo, wu2024beta, xie2024v, xiong2024llava, yang2025mitigating}. 

Standard DPO methods treat the reasoning process as a single structure. To address this, preference granularity in DPO has been explored at the token~\cite{liu2024tis,shao2025earlier,zeng2024token,zhao2024epo,zhong2024dpo}, step~\cite{lai2024step, zhang2024chain}, sentence~\cite{pang2024iterative,rafailov2023direct,she2024mapo}, and turn~\cite{shi2024direct,song2024trial,xiong2409building} levels.
While effective in certain domains, these approaches overlook the semantic roles of different segments in LongCoT, where descriptive grounding and logical reasoning require distinct optimization.
In contrast, our proposed \dponame{} introduces functional-level preference granularity.\looseness-1

\textbf{Multi-LLM Guided Reasoning}
Recent work has explored leveraging multiple LLMs to collaboratively solve complex reasoning tasks, often integrated with Monte Carlo Tree Search (MCTS). Methods such as MoA \cite{wang2024mixture}, MoSA \cite{yang2025multi}, AlphaLLM-CPL \cite{wang2024towards}, and LE-MCTS \cite{park2024ensembling} enhance multi-agent text-based reasoning using ensemble methods and stepwise search. CoMCTS (Mulberry) \cite{yao2024mulberry} extends multi-LLM MCTS to multimodal reasoning, primarily targeting collaborative reflection and error correction. In contrast, our method, \mctsname{}, addresses the challenge of spatial reasoning in VLMs, introducing fine-grained preference learning and multi-level spatial rewards that allow for coherent, visually-grounded reasoning paths across multimodal data.

%% file: sections/03_method.tex
\section{Method}
\label{sec:Methodology}
\input{assets/figures/overviewfig}

\subsection{Spatial Reasoning from Images}
\label{sec:spatialsetting}
Spatial reasoning is a core vision-language challenge, requiring models to understand visual layouts and perform logical inference over spatial relationships. We define spatial reasoning as a multimodal understanding problem where the goal is to generate accurate reasoning paths based on visual and textual inputs. Formally, a spatial reasoning instance can be represented as a tuple 
$\mathcal{T}\!=\!(\bm{I}, \bm{Q}, \bm{P}) \xrightarrow{\pi_{\theta}} \bm{R}$
where  $\bm{I}$ represents the input image containing the visual content, $\bm{Q}$ is the textual query specifying the spatial reasoning task, $\bm{P}$ denotes the visual prompt tokens pointing to a specific object or region in the image, and $\bm{R}$ the textual response, providing the answer or step-by-step reasoning path. The primary objective of a spatial reasoning model, denoted as $\pi_{\theta}$, is to map the multimodal input $\mathcal{T}$ to a logically sound and spatially grounded response $\bm{R}$.

\input{assets/figures/architecture}
Unlike typical direct-response VQA tasks, \textbf{\modelname{}} is designed to output LongCoT reasoning traces that decompose spatial reasoning into clear, verifiable steps. 
To train the model, we introduce a fine-grained preference objective that optimizes descriptive and reasoning responses separately (\Cref{sec:f-dpo}) and a spatial reward mechanism that evaluates candidate reasoning paths based on spatial and logical understanding (\Cref{sec:rlaif}). Finally, to address the lack of LongCoT supervision for spatial reasoning, we propose a multi-model collaborative tree search method that generates diverse, reward-aligned reasoning trajectories to enable preference-based training (\Cref{sec:mmcts}).
An overview of the proposed training framework is depicted in Figure~\ref{fig:totalstructure} and the \modelnamenc{} architecture is shown in Figure~\ref{fig:architecture}.

\subsection{Fine-grained Direct Preference Optimization (\dponame)}
\label{sec:f-dpo}

We propose \dponame{} as a novel fine-grained off-policy preference learning algorithm to optimize LongCoT spatial reasoning. Traditional DPO methods apply a single global trade-off parameter $\beta$ uniformly across all reasoning steps~\cite{wang2024mdpo, wu2024beta,yang2025mitigating}, implicitly treating all response segments as equally learnable. 
However, this can lead to degenerate solutions, as the model may overfit to simpler descriptive responses while under-optimizing the more complex reasoning paths. This observation motivates the design of our fine-grained preference mechanism, which introduces segment-level preference granularity.

To facilitate fine-grained preference optimization, we first segment each LongCoT response $\bm{R}$ into its constituent description $\bm{R}_{\text{desc}}$ and reasoning $\bm{R}_{\text{reason}}$ components, represented as $\bm{R}\!=\![\bm{R}_{\text{desc}}, \bm{R}_{\text{reason}}]$. We then quantify the preference signal for each segment by calculating the score difference between the corresponding segments derived from the positive $\bm{R}^{p}$ and negative responses $\bm{R}^{l}$, yielding segment-wise preference differentials:
\begin{equation}\label{eq:reward}
\Delta \bm{R}_{\text{desc}} = \operatorname{score}\bigl(\bm{R}^{p}_{\text{desc}}\bigr) - \operatorname{score}\bigl(\bm{R}^{l}_{\text{desc}}\bigr), \quad \Delta \bm{R}_{\text{reason}} = \operatorname{score}\bigl(\bm{R}^{p}_{\text{reason}}\bigr) - \operatorname{score}\bigl(\bm{R}^{l}_{\text{reason}}\bigr),
\end{equation}
where the differentials $\Delta \bm{R}_{\text{desc}}$ and $\Delta \bm{R}_{\text{reason}}$ quantify the preference margin for description and reasoning segments based on the preference pair, and $score(\cdot)$ composite scores are introduced in Section \ref{sec:rlaif}. The design of \dponame{} is guided by two key principles:

\vspace{.1cm}
\textbf{Principle 1}: \textit{Preference optimization strength should be dynamically balanced according to the intrinsic complexity and quality disparity between description and reasoning components.}

Our analysis of the fine-grained reward signals reveals that descriptive segments ($\bm{R}_{\text{desc}}$) are easier to optimize while models struggle with reasoning segments ($\bm{R}_{\text{reason}}$) that are typically longer and require multi-hop logic. Thus, a unified optimization parameter $\beta$ may lead to reasoning under-optimization. To address this, \dponame{} introduces separate, adaptively-tuned trade-off parameters, $\beta_{\text{desc}}$ and $\beta_{\text{reason}}$, which dynamically control the learning signals for each segment independently, to allow the model to prioritize deeper logical inference while maintaining visual and attribute accuracy.

\vspace{.1cm}
\textbf{Principle 2}: \textit{The choice of segment-specific optimization parameters ($\beta_{\text{desc}}$ and $\beta_{\text{reason}}$) should prioritize the component that exhibits a larger preference differential, such that learning focuses on harder-to-learn segments.}

We further empirically observe that the preference score differential for descriptive components $\Delta \bm{R}_{\text{desc}}$ is consistently smaller than $\Delta \bm{R}_{\text{reason}}$. To account for this, fDPO computes dynamic segment weights $w_{\text{desc}}$ and $w_{\text{reason}}$ to adaptively adjust the learning signals for the description and reasoning components, respectively:
\begin{equation}\label{eq:w}
w_{s}
= \frac{\exp\bigl(\lambda\, \cdot \Delta \bm{R}_{s}\bigr)}
         {\exp\bigl(\lambda \cdot \Delta \bm{R}_{\text{desc}}\bigr) + \exp\bigl(\lambda \cdot \Delta \bm{R}_{\text{reason}}\bigr)},
\quad
s\in\{\text{desc},\text{reason}\},
\end{equation}
where $\Delta \bm{R}_{s}$ is the preference score differential for segment $s$ (either description or reasoning), $\lambda>0$ controls the sensitivity of weights, and $\{w_{\text{desc}},w_{\text{reason}}\}$ reflect the relative importance of each segment.

These weights are then mapped to adjustment factors centered around 1 and  applied to the base optimization parameter $\beta$ to yield segment-specific trade-off parameters $\beta_{\text{desc}}$ and $\beta_{\text{reason}}$ for description and reasoning, respectively:
\begin{equation}\label{eq:beta_scale}
\beta_{s}
= \beta\ \times f(w_{s})
= \beta\bigl[1 + \alpha\,(2w_{s}-1)\bigr],
\quad
s\in\{\text{desc},\text{reason}\},
\end{equation}
where $w_s \in [0,1]$ is the respective segment-specific weight (either description or reasoning), $\alpha$ is a hyperparameter that controls the maximum scaling amplitude, and $\beta$ is the base hyperparameter value. This design implements a dynamic learning strategy: segments with larger preference differentials (higher relative importance $w_s$) receive a higher effective $\beta_{s}$, amplifying the learning signal and prioritizing those components. Conversely, smaller preference differentials yield lower $\beta_{s}$, enabling finer-grained updates. This adaptive mechanism allows fDPO to balance optimization based on segment-specific learning difficulty, promoting better alignment for complex reasoning steps while preserving descriptive accuracy.
The final optimization objective for each segment is defined as 
\begin{equation}\label{eq:fdpo}
\mathcal{F}_{s}(\mathcal{T},\bm{R}^p,\bm{R}^l)
=\,
\log\frac{\pi_\theta\bigl(\bm{R}^p_{s}\mid \mathcal{T}\bigr)}
         {\pi_{\text{ref}}\bigl(\bm{R}^p_{s}\mid \mathcal{T}\bigr)}
\;-\;
\log\frac{\pi_\theta\bigl(\bm{R}^l_{s}\mid \mathcal{T}\bigr)}
         {\pi_{\text{ref}}\bigl(\bm{R}^l_{s}\mid \mathcal{T}\bigr)},
\quad
s\in\{\text{desc},\,\text{reason}\}.
\end{equation}
Here, $\mathcal{T}$ represents the multimodal input (image, text query, visual prompt), $\pi_{\theta}$ is the model's learned policy, and $\mathcal{F}_s$ measures the segment-specific preference margin in log-likelihood ratios relative to a reference policy $\pi_{\text{ref}}$. The overall optimization objective for \dponame{} is
\begin{equation}\label{eq:L_FDPO}
\scalebox{0.99}{$
\mathcal{L}_{\text{fDPO}}(\theta)
= -\mathbb{E}_{(\mathcal{T},\bm{R}^p,\bm{R}^l)\sim\mathcal{D}}
\Biggl[\,
\log \sigma\!\Bigl(
\beta_{\text{desc}}\;\mathcal{F}_{\text{desc}}(\mathcal{T},\bm{R}^p,\bm{R}^l)
\;+\;
\beta_{\text{reason}}\;\mathcal{F}_{\text{reason}}(\mathcal{T},\bm{R}^p,\bm{R}^l)
\Bigr)\Biggr],
$}
\end{equation}
where $\mathcal{D}\!=\!\{ (\mathcal{T}^{(i)}, \bm{R}^{(p,i)}, \bm{R}^{(l,i)})\}_{i=1}^N$ and $\sigma(\cdot)$ sigmoid activation function.

\input{assets/figures/dpofig}

\subsection{Fine-Grained Spatial Rewards}
\label{sec:rlaif}

To optimize spatial reasoning paths effectively, we introduce a fine-grained reward mechanism that evaluates candidate reasoning paths across visual, spatial, and logical dimensions. 
Rewards capture alignment with image content, spatial relationships, and logical inference. Figure~\ref{fig:fdpo} illustrates the proposed fine-grained spatial rewards for \dponame{}. Specifically, we define four scalar rewards; details about their formulations and rationale behind each reward are provided in Appendix~\ref{appendix:rewards}.

\begin{itemize}[itemsep=0.5ex, parsep=0pt, topsep=0pt, leftmargin=0.5cm,label={\Large$\diamond$}]
\item \textbf{Visual Consistency Reward (\(\mathcal{R}_{\text{vc}}\))} evaluates the description $\bm{R}_{\text{desc}}$ to ensure spatial grounding and fidelity. The reward verifies key aspects of the quality and alignment of the description with the visual scene, such as whether all referenced objects are present and identifiable, whether the stated properties (such as color, size, and shape) match the visual content, whether the description includes all necessary details prompted by the query, and whether it remains contextually appropriate and free from extraneous information.

\item \textbf{Depth-Guided Spatial Reward (\(\mathcal{R}_{\text{sp}}\))} measures fine-grained spatial understanding by leveraging depth information. This reward is independently computed for the description $\bm{R}_{\text{desc}}$ and reasoning $\bm{R}_{\text{reason}}$ components, with two adaptive weighting mechanisms: an uncertainty weight that adjusts the score for spatial expressions with qualifiers (\eg"approximately","possibly") to account for reduced confidence, and a context-aware weight that emphasizes spatial relations directly relevant to the query. 
The final spatial rewards are computed as the uncertainty and context-aware weighted correctness scores across all spatial assertions in the corresponding description and reasoning components, validated against both the RGB image and its corresponding depth map. This ensures that more confident and contextually aligned relations have stronger influence on the reward.\looseness-1

\item \textbf{Logical Coherence Reward
(\(\mathcal{R}_{\text{lc}}\))} evaluates the reasoning $\bm{R}_{\text{reason}}$ for structural integrity and logical correctness. This reward captures multi-hop inference and factual alignment by verifying that premises are consistent with the image, depth map, and preceding descriptions, reasoning steps maintain spatial and causal logic, the application of physical, spatial, and logical principles remains coherent throughout, and the conclusion is fully supported by the reasoning chain.\looseness-1
\end{itemize}

The preference differentials in \dponame{} are computed from composite rewards that aggregate fine-grained evaluations across segments: $\operatorname{score}(\bm{R}_{\text{desc}}) \!=\!\mathcal{R}_{\text{vc}} + \mathcal{R}_{\text{sp,desc}}$ and $\operatorname{score}(\bm{R}_{\text{reason}})\!= \!\mathcal{R}_{\text{lc}} + \mathcal{R}_{\text{sp,reason}}$.

\subsection{Multi-Model MCTS (\mctsname)}\label{sec:mmcts}
We introduce a Multi-Model Monte Carlo Tree Search (\mctsname) framework for generating high-quality LongCoT data $\mathcal{D}\!=\!\{(\mathcal{T},\bm{R}^p,\bm{R}^l)\}_{i=1}^N$ tailored to spatial reasoning. Inspired by DeepSeek-R1-Zero~\cite{deepseekr1zero2024} and prior multimodal MCTS methods~\cite{yao2024mulberry,park2024ensembling,yang2025multi}, \mctsname{} explores diverse reasoning trajectories across multiple VLMs to effectively search for logical, spatially-consistent explanations that satisfy the query. Formally, the reasoning process is defined as a sequence of reasoning states:
$\bm{S}\!=\!\{s_0, \dots, s_t, \dots, s_T\}$, where $s_t \in \bm{S}$ represents a partial reasoning state, $s_0$ is the initial state derived from $\mathcal{T}$, and $s_T$ is a terminal state corresponding to a fully reasoned path.
\mctsname{} operates through four key stages: \textit{Expand, Simulate, Backprop,} and \textit{Select}. 

\textbf{Expand.} At each step $t$, \mctsname{} expands the current state \(s_t\) by generating  diverse candidate reasoning states $\mathcal{S}_{\text{c}}$ using multiple VLMs $\{\pi_k\}_{k=1}^{K}$ concurrently, \ie
\begin{equation}
\mathcal{S}_{\text{c}} = \bigcup_{k=1}^K \pi_k\Big(s_t \mid \mathcal{T}, \text{Parent}(s_t)\Big),
\end{equation}
where \(\pi_k\) is the $k$-th VLM,  \(\mathcal{T}\) multimodal input, and \(\text{Parent}(s_t)\) ancestor reasoning states of \(s_t\). To ensure consistency, we enforce a structured output format across all VLMs (Appendix~\ref{appendix:MMCTS_format}).

\textbf{Simulate.} Each candidate \(s_{k,t} \in \mathcal{S}_{\text{c}}\) generated during expansion is evaluated based on three distinct criteria: \textbf{(i) visual} description accuracy against the original image, \textbf{(ii) spatial} correctness of inferred spatial relationships utilizing both original and depth-derived images, and \textbf{(iii) logical} coherence of the textual reasoning steps. 
The evaluation score \(R(s_{k,t})\) is computed as:
\begin{equation}
R(s_{k,t}) = \frac{1}{M}\sum_{m=1}^{M}\left[\mathbb{I}^{(m)}_{\text{visual}}(s_{k,t}) + \mathbb{I}^{(m)}_{\text{spatial}}(s_{k,t}) + \mathbb{I}^{(m)}_{\text{logical}}(s_{k,t})\right],
\end{equation}
where \(M\) is the number of evaluation models, and each \(\mathbb{I}^{(m)}_{\text{eval}}(s_{k,t})\) indicator function defined as:
\begin{equation}
\mathbb{I}^{(m)}_{\text{eval}}(s_{k,t}) = \{+1\ (\text{fully accurate}),\ 0\ (\text{neutral}),\ -1\ (\text{inaccurate})\}.
\end{equation}

We preserve high-quality paths by pruning the candidate set according to the evaluation score, \ie
$\mathcal{S}_{\text{c}}^*\!=\!\{ s_{k,t} \mid R(s_{k,t}) \geq 0 \}.$ Appendix~\ref{appendix:MMCTS_evaluation} provides detailed descriptions of the evaluation. 

\textbf{Backprop.} To perform credit assignment, scores from the simulation phase are recursively propagated upwards through the search tree. The objective is to update the value estimates \( V(s_{k,t}) \) and visit counts \( N(s_{k,t}) \) for each parent node \(s_{k,t}\) based on the performance of its children $\mathcal{S}_{\text{c}}^*\!=\!\text{Child}(s_{k,t})$,
\begin{equation}
        V(s_{k,t}) \leftarrow \frac{
            N(s_{k,t}) V(s_{k,t}) + \sum_{s_c \in \mathcal{S}_{\text{c}}^*} N(s_c) R(s_c)
        }{
            N(s_{k,t}) + \sum_{s_c \in  \mathcal{S}_{\text{c}}^*} N(s_c)
        }, \quad
        N(s_{k,t}) \leftarrow N(s_{k,t}) + \sum_{s_c \in  \mathcal{S}_{\text{c}}^*} N(s_c).
\end{equation}

\textbf{Select.} This phase is responsible for choosing the most promising candidate state for further exploration in the next iteration of tree expansion. We use the Upper Confidence Bound (UCB) strategy to select the next state  $s_{k',t+1}^{\star}$ to traverse, based on updated values and visitation statistics. UCB ensures that high-value paths are prioritized, while also exploring less-visited nodes to discover new reasoning trajectories. The candidate selected maximizes the UCB objective, \ie
\begin{equation}
s_{k',t+1}^{\star}
\;=\;
\operatorname*{arg\,max}_{s_c \in \mathcal{S}_{\text{c}}^*}
\left[
        V(s_c)\;+\;
        \tilde{\alpha}\,
        \sqrt{\frac{\log N(s_{k,t})}{1 + N(s_c)}}~
\right],
\end{equation}
where \(V(s_c)\)  is the value estimate of the candidate state $s_c$, \(N(s_c)\) its visit count, and \(\tilde{\alpha}>0\) is a hyperparameter that balances exploration \vs exploitation.

%% file: assets/figures/overviewfig.tex
\begin{figure}[!t]
    \centering    
    \includegraphics[width=\linewidth]{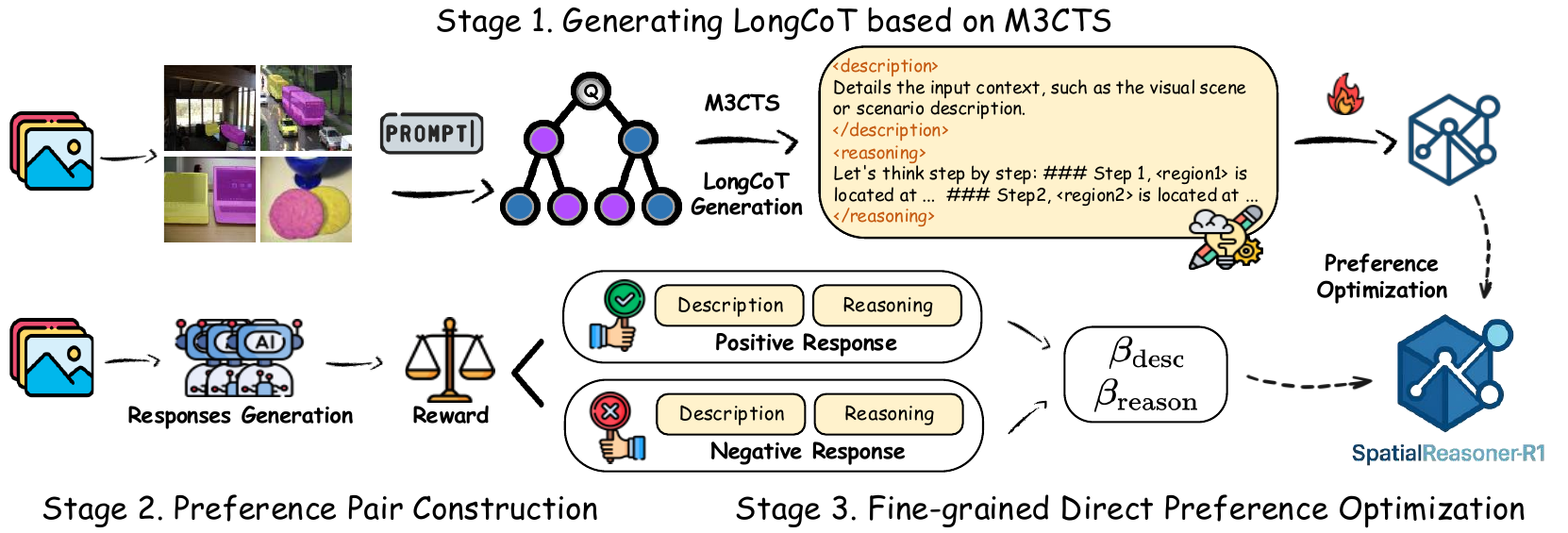}
    \caption{\textbf{Method Overview.} To train \modelnamenc{}, we (1) generate reasoning paths using \mctsname{}, (2) construct fine-grained preference pairs via reward-based selection, and (3) train with fine-grained DPO (\dponame{}) to optimize descriptive and logical reasoning separately.}
    \label{fig:totalstructure}
\end{figure}

%% file: assets/figures/architecture.tex
\begin{wrapfigure}{r}{0.65\textwidth}
    \centering
    \includegraphics[width=0.99\linewidth]{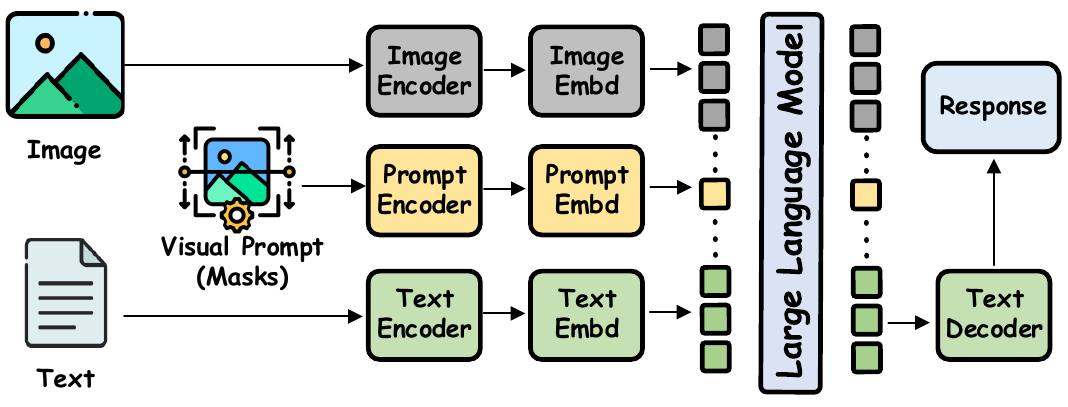}
    \vspace{-0.2cm}
    \caption{\textbf{Architecture Overview.} \modelnamenc{} is a VLM that takes as input a text instruction, visual prompts, and an image, and generates LongCoT reasoning responses.}
    \label{fig:architecture}
    \vspace{-3mm}
\end{wrapfigure}

%% file: assets/figures/dpofig.tex
\begin{figure}[!t]
    \centering
    \includegraphics[width=0.99\linewidth]{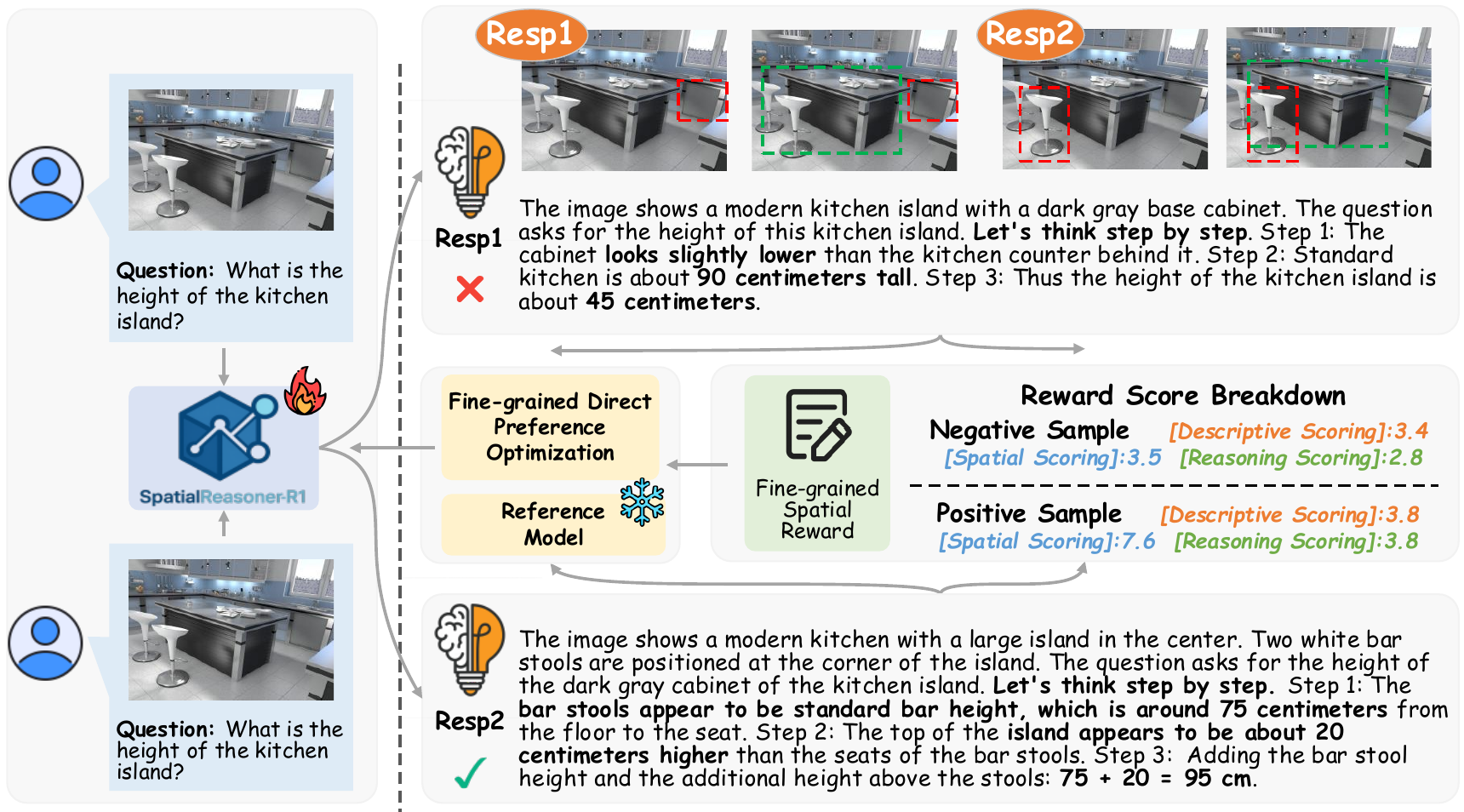}
    \vspace{-0.2cm}
    \caption{\textbf{Fine-Grained Spatial Rewards.}
    Candidate reasoning paths are decomposed into three aspects, \emph{descriptive}, \emph{spatial}, and \emph{reasoning}, scored separately; the higher value in each row is marked by
    \raisebox{-0.2ex}{\includegraphics[height=0.8em]{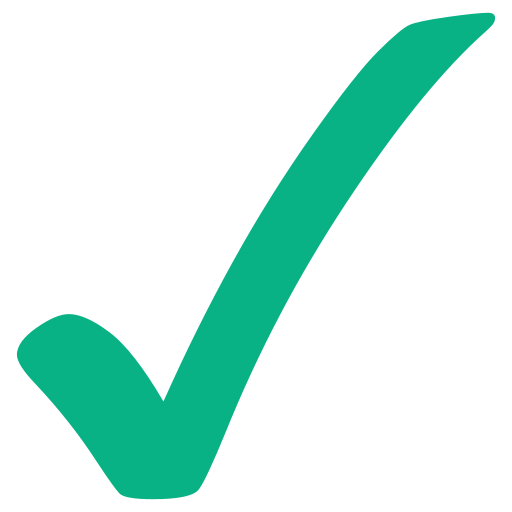}}
    and the lower by
    \raisebox{-0.2ex}{\includegraphics[height=0.8em]{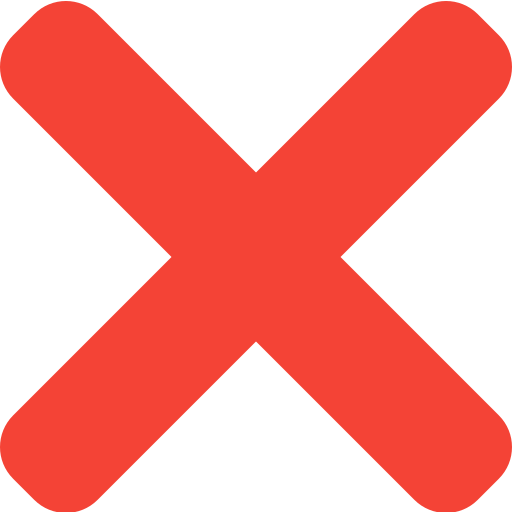}}.
    \textbf{Explanation of Scoring:}
    \emph{Descriptive:} Negative response omits the two bar-stools and uses generic “modern kitchen” wording, whereas the positive response lists every salient object; 
    \emph{Spatial:} Negative response wrongly claims the island is \emph{lower} than the rear counter and ignores the 20cm offset revealed by the stool reference, whereas the positive response provides its estimate to the 75cm stool height plus that offset; 
    \emph{Reasoning:} Negative response uses an illogical ``half-height'' heuristic \(90\text{cm} \rightarrow 45\text{cm}\) without intermediate computation, whereas the positive response explicitly adds reference height and gap (75cm + 20cm = 95cm).  
    These per-category deficits yield lower composite reward, designating the upper response as negative sample.}
    \label{fig:fdpo}
\end{figure}

%% file: sections/04_experiments.tex
\section{Experiments}
\label{sec:experiments}

\subsection{Experimental Setup}
We evaluate \modelnamenc{} across diverse spatial reasoning and general vision-language established benchmarks to assess the model's fine-grained spatial understanding and logical reasoning capabilities. Implementation details are provided in Appendix \ref{appendix:training}.

\textbf{Spatial Reasoning Benchmarks.} Our primary benchmark is \textsc{SpatialRGPT-Bench}~\cite{cheng2024spatialrgpt}, comprising image-based spatial reasoning questions and their corresponding ground truth answers. Detailed descriptions of benchmarks and evaluation protocols are provided in Appendix \ref{appendix:benchmark}.

\input{assets/tables/spatial_ability}

\textbf{General Vision-Language Benchmarks.} To validate the robustness of \modelnamenc{} beyond purely spatial tasks, we evaluate on broader vision-language datasets such as \textsc{MME}, \textsc{POPE}, \textsc{SEED-Bench}, \textsc{AI2D}, \textsc{SQA-test}, \textsc{MMMUv}, \textsc{MMStar}, and \textsc{HallusionBench}~\cite{liang2024survey, li2023evaluating, li2023seed, kembhavi2016diagram, lu2022learn, yue2024mmmu, chen2024we, guan2024hallusionbench}.
These datasets cover fundamental vision-language tasks such as object grounding, hierarchical scene parsing, multimodal understanding, and multi-turn reasoning in diverse multimodal contexts.

\textbf{Baselines.}
We benchmark \modelnamenc{} against two categories of baseline models:\\
\noindent \tightcolorbox{genLLM}{\textbf{General Large VLMs.}} This includes powerful, widely-accessible models such as  Gemini 2.0 Flash~\cite{google-deepmind-gemini2024}, Llama 4 Maverick~\cite{meta-llama4-2025}, Gemini 1.5 Pro~\cite{team2024gemini}, and ChatGPT-4o~\cite{hurst2024gpt}. These are evaluated in zero-shot/few-shot settings as reference of standard VLM capabilities w/o task-specific fine-tuning.\\
\noindent \tightcolorbox{specLLM}{\textbf{Specialized VLMs.}} This baseline set comprises models specifically developed, adapted, or fine-tuned for spatial understanding tasks, allowing us to assess our contributions relative to other specialized approaches. The models included are: SpatialBot-3B~\cite{cai2024spatialbot}, SpaceThinker Qwen2.5VL-3B~\cite{bai2025qwen2}, InternVL2.5-78B~\cite{chen2024internvl}, Sa2VA (4B, 8B)~\cite{yuan2025sa2va}, and SpatialRGPT-8B~\cite{cheng2024spatialrgpt}.\\
\tightcolorbox{ourVLM}{\textbf{\modelname{} VLMs.}}
We also include \modelnamenc{} 4B and 8B variants with different training strategies, such as \modelnamenc{} SFT,  \modelnamenc{} DPO, trained with standard DPO, and \modelnamenc{} \dponame{} trained with the proposed fine-grained DPO method.

\subsection{Experimental Results}
\label{sec:exp_results}
\textbf{Spatial Reasoning.}
As shown in Table~\ref{tab:spatial_ability}, \modelnamenc{} models achieve substantial improvements over both general-purpose and spatial-specialized VLMs across all spatial tasks. Notably, \modelnamenc{} \dponame{} 8B sets a new benchmark for average accuracy with \textbf{2.9\%} and \textbf{15.8\%} gains over SpatialRGPT-8B on spatial qualitative and quantitative tasks, respectively. Our parameter-efficient \modelnamenc{} \dponame{} 4B outperforms larger models like InternVL2.5-78B, highlighting the effectiveness of our fine-tuning strategy. 
Finally, when compared to its predecessor DPO 8B, our optimized variant \dponame{} 8B boosts average accuracy by \textbf{4.1\%} across qualitative tasks and by \textbf{9.0\%} in quantitative tasks.\looseness-1

\textbf{General Vision-Language Understanding.} Beyond achieving state-of-the-art performance in spatial reasoning tasks, our \modelnamenc{} \dponame{} 8B also demonstrates significant gains in general vision-language benchmarks compared to SpatialRGPT-8B, as presented in Table~\ref{tab:general_ability}. 

\input{assets/tables/general_ability}
\input{assets/figures/qualfig}

\subsection{Qualitative Examples}
Figure \ref{fig:qual} provides qualitative examples that demonstrate \modelnamenc{}'s advanced capability for coherent, multi-step spatial reasoning. \textbf{\modelname{}} first estimates the fireplace and TV-stand widths at 1.2m and 1.4m, then computes
\( \frac{1.2}{2} + 0.2\,\text{(gap)} + \frac{1.4}{2} = 1.5\text{m} \),
a value that nearly matches the reference while transparently tying each term to an observed feature.  
In contrast, \textbf{InternVL2.5-78B} adopts similar width guesses (1.2m, 1.5m) but assumes ``the distance between the fireplace and the TV stand seems to be about the width of one object''. This assumption is inconsistent with what is shown in the image. \textbf{Gemini 1.5 Pro} aligns the fireplace’s right edge with the TV’s left edge, assigns both objects a 1m width, and combines only half a fireplace width (0.5m) with one quarter of the stand width (0.25m). These two estimates are not accurate and ignore the gap between the two regions. \textbf{SpatialRGPT-8B} yields a more accurate estimate than Gemini 1.5 Pro and InternVL2.5-78B. However, since it is not designed as a reasoning model, it cannot generate step-by-step reasoning traces, \ie does not explicitly reveal the logical chain of spatial deductions or intermediate calculations leading to that estimate. These qualitative examples show that \modelnamenc{} has more accurate spatial awareness. Additional examples can be found in Appendix \ref{appendix:examples}.\looseness-1

\subsection{Ablations} 
Table~\ref{tab:alpha_compact} illustrates the impact of varying the Alpha ($\alpha$) parameter, which modulates the magnitude of segment-specific learning adjustments during \dponame{} optimization. When $\alpha$ is set too high, the model may overly focus on the reasoning part at the expense of the other, introducing instability and degraded performance, as observed when $\alpha$ reaches 40\%. Conversely, if $\alpha$ is too low, both description and reasoning segments are optimized equally. A moderate value of $\alpha=30\%$ allows the model to effectively amplify learning signals for fine-grained spatial distinctions, leading to substantial improvements across all spatial metrics. 
Furthermore, Table~\ref{tab:lambda_compact} presents the impact of varying the Lambda ($\lambda$) parameter, which modulates the sensitivity of segment-specific weights to preference differentials, controlling how responsively the model shifts learning focus based on the observed preference margins. As $\lambda$ increases, the model becomes more sensitive to segment-specific preference differences, leading to noticeable changes in performance across spatial metrics.
We observe that while a moderate value of $\lambda=0.6$ achieves the best overall results, setting $\lambda$ too high can introduce slight performance degradation in some spatial metrics, likely due to overly aggressive re-weighting. 
\input{assets/tables/alpha_lambda_compact}

%% file: assets/tables/spatial_ability.tex
\begin{table}[t!]
  \caption{\textbf{Spatial Reasoning Success Rates ($\uparrow$) on \textsc{SpatialRGPT-Bench}}. Classification (top) and numeric distance/direction (bottom).
    \colorbox{genLLM}{\phantom{\rule{1ex}{1ex}}} are General Large VLMs,
    \colorbox{specLLM}{\phantom{\rule{1ex}{1ex}}} are Customized VLMs,
    \colorbox{ourVLM}{\phantom{\rule{1ex}{1ex}}} are \textbf{\modelname{}} variants.
    ``/'' indicates the model refuses to provide a response for that metric.}
  \label{tab:spatial_ability}
  \vspace{-0.2cm}
  \centering
  {\fontsize{8pt}{11pt}\selectfont
   \resizebox{\linewidth}{!}{%
     \begin{NiceTabular}{l c c c c c c c}[colortbl-like]
       \toprule[1.2pt]
       %\rowcolor{gray!20}
         & \begin{tabular}{c}\textbf{Below}/\\\textbf{Above}\end{tabular}
         & \begin{tabular}{c}\textbf{Left}/\\\textbf{Right}\end{tabular}
         & \begin{tabular}{c}\textbf{Big}/\\\textbf{Small}\end{tabular}
         & \begin{tabular}{c}\textbf{Tall}/\\\textbf{Short}\end{tabular}
         & \begin{tabular}{c}\textbf{Wide}/\\\textbf{Thin}\end{tabular}
         & \begin{tabular}{c}\textbf{Behind}/\\\textbf{Front}\end{tabular}
         & \begin{tabular}{c}\textbf{Qual.}\\\textbf{Acc.}\end{tabular} \\
       \midrule[1.2pt]\midrule 

           \rowcolor{genLLM}Gemini 2.0 Flash \cite{google-deepmind-gemini2024}   & 58.33 & 68.57 & 16.98 & 50.00 & 15.38 & 53.63 & 44.29 \\
           \rowcolor{genLLM}Llama 4 Maverick \cite{meta-llama4-2025}   & 54.17 & 61.90 & 33.02 & 50.89 & 25.96 & 55.45 & 47.18 \\
           \rowcolor{genLLM}Gemini 1.5 Pro \cite{team2024gemini}     & 85.83 & 56.19 & 58.49 & 71.42 & 55.76 & 60.00 & 65.14 \\
           \rowcolor{genLLM}ChatGPT-4o \cite{hurst2024gpt}         & 87.50 & 80.00 & 53.77 & 63.39 & 51.92 & 60.90 & 66.67 \\
           \midrule 

           \rowcolor{specLLM}SpatialBot-3B \cite{cai2024spatialbot}     & 52.50 & 62.86 & 57.54 & 49.11 & 49.04 & 62.73 & 55.56 \\
           \rowcolor{specLLM}SpaceThinker Qwen2.5VL-3B \cite{bai2025qwen2} & 89.16 & 63.81 & 76.41 & 56.25 & 56.73 & 70.91 & 69.25 \\
           \rowcolor{specLLM}InternVL2.5-78B \cite{chen2024internvl}   & 94.16 & 94.28 & 64.15 & 65.17 & 55.76 & 58.18 & 72.29 \\
           \rowcolor{specLLM}Sa2VA 4B \cite{yuan2025sa2va}          & 22.50 & 25.71 & 25.47 & 16.07 & 27.88 & 30.91 & 24.65 \\
           \rowcolor{specLLM}Sa2VA 8B \cite{yuan2025sa2va}          & 50.00 & 39.04 & 45.28 & 26.78 & 45.19 & 53.63 & 43.37 \\
           \rowcolor{specLLM}SpatialRGPT-8B \cite{cheng2024spatialrgpt}    & \textbf{99.17} & \textbf{100.00} & 84.90 & 89.28 & \textbf{91.34} & 90.90 & \underline{92.69} \\
           \midrule 
           
           \rowcolor{ourVLM}\textbf{\modelname{}} SFT 4B         & 79.16 & 78.09 & 55.66 & 66.96 & 59.61 & 75.45 & 69.41 \\
           \rowcolor{ourVLM}\textbf{\modelname{}} SFT 8B         & 81.66 & 81.90 & 75.47 & 75.89 & 79.80 & 83.63 & 79.75 \\
           \rowcolor{ourVLM}\textbf{\modelname{}} DPO 4B         & 91.66 & 91.42 & 69.81 & 65.17 & 71.15 & 85.45 & 79.29 \\
           \rowcolor{ourVLM}\textbf{\modelname{}} DPO 8B         & 94.16 & 93.33 & \underline{89.62} & \underline{90.18} & \underline{88.64} & \underline{92.27} & 91.48 \\  
           \rowcolor{ourVLM}\textbf{\modelname{}} \textbf{\dponame{}} 4B       & 95.83 & 93.33 & 83.96 & 74.10 & 87.50 & 89.09 & 87.37 \\
           \rowcolor{ourVLM}\textbf{\modelname{}} \textbf{\dponame{}} 8B       & \underline{98.33} & \underline{98.10} & \textbf{95.28} & \textbf{96.43} & \textbf{91.34} & \textbf{93.64} & \textbf{95.59} \\
           \bottomrule
         \end{NiceTabular}
   }
  }
  
  \vspace{3pt} 

  {\fontsize{8pt}{11pt}\selectfont
   \resizebox{\linewidth}{!}{%
     \begin{NiceTabular}{l c c c c c c c}[colortbl-like] 
       \toprule[1.2pt]
       %\rowcolor{gray!20}
         & \begin{tabular}{c}\textbf{Direct}\\\textbf{Distance}\end{tabular}
         & \begin{tabular}{c}\textbf{Horizontal}\\\textbf{Distance}\end{tabular}
         & \begin{tabular}{c}\textbf{Vertical}\\\textbf{Distance}\end{tabular}
         & \begin{tabular}{c}\textbf{Width}\end{tabular}
         & \begin{tabular}{c}\textbf{Height}\end{tabular}
         & \begin{tabular}{c}\textbf{Direction}\end{tabular}
         & \begin{tabular}{c}\textbf{Quan.}\\\textbf{Acc.}\end{tabular} \\
       \midrule[1.2pt]\midrule 

       \rowcolor{genLLM}Gemini 2.0 Flash \cite{google-deepmind-gemini2024}   &  9.45 & 10.65 & 26.41 & 10.52 & 30.82 & 54.20 & 22.43 \\
       \rowcolor{genLLM}Llama 4 Maverick \cite{meta-llama4-2025}   & 24.48 & 28.68 & 34.28 & 35.71 & 44.61 & 58.09 & 36.72 \\
       \rowcolor{genLLM}Gemini 1.5 Pro \cite{team2024gemini}     & 14.18 & 17.21 & 14.15 & 19.54 & 36.09 & 30.84 & 21.90 \\
       \rowcolor{genLLM}ChatGPT-4o \cite{hurst2024gpt}         & / & / & / & / & / & 60.75 & / \\
       \midrule

       \rowcolor{specLLM}SpatialBot-3B \cite{cai2024spatialbot}     &  6.00 & 15.51 &  8.00 & 10.52 & 18.75 & 39.00 & 15.62 \\
       \rowcolor{specLLM}SpaceThinker Qwen2.5VL-3B \cite{bai2025qwen2} & 24.32 & 17.21 & 59.43 & 23.27 & 23.62 & 32.35 & 28.97 \\
       \rowcolor{specLLM}InternVL2.5-78B \cite{chen2024internvl}   & 27.70 & 22.13 & 41.50 & 29.32 & 34.58 & 62.61 & 35.25 \\
       \rowcolor{specLLM}Sa2VA 4B \cite{yuan2025sa2va}          & 13.51 & 15.57 & 19.81 & 13.53 & 12.03 & 10.28 & 14.02 \\
       \rowcolor{specLLM}Sa2VA 8B \cite{yuan2025sa2va}          & 14.18 & 14.75 &  9.43 & 14.28 & 19.54 & 14.18 & 14.55 \\
       \rowcolor{specLLM}SpatialRGPT-8B \cite{cheng2024spatialrgpt}    & 45.90 & \underline{68.00} & 56.60 & 48.90 & 61.70 & \textbf{95.30} & 61.42 \\
       \midrule 

       \rowcolor{ourVLM}\textbf{\modelname{}} SFT 4B         & 22.29 & 27.86 & 31.13 & 25.56 & 33.80 & 47.66 & 30.71 \\
       \rowcolor{ourVLM}\textbf{\modelname{}} SFT 8B         & 28.43 & 20.49 & 44.05 & 33.59 & 51.63 & 46.72 & 37.12 \\
       \rowcolor{ourVLM}\textbf{\modelname{}} DPO 4B         & 47.97 & 46.72 & 60.37 & 45.11 & 55.63 & 91.58 & 56.61 \\
       \rowcolor{ourVLM}\textbf{\modelname{}} DPO 8B         & \underline{62.83} & 56.55 & 60.37 & \underline{70.45} & \underline{68.42} & 93.45 & \underline{68.22} \\
       \rowcolor{ourVLM}\textbf{\modelname{}} \textbf{\dponame{}} 4B       & 60.13 & 59.01 & \underline{71.70} & 65.41 & 57.89 & 92.52 & 66.76 \\
       \rowcolor{ourVLM}\textbf{\modelname{}} \textbf{\dponame{}} 8B       & \textbf{70.95} & \textbf{72.13} & \textbf{74.52} & \textbf{80.45} & \textbf{74.43} & \underline{94.39} & \textbf{77.30} \\

       \bottomrule
     \end{NiceTabular}
   }
  }
  \vspace{-0.5cm}
\end{table}

%% file: assets/tables/general_ability.tex
\begin{table}[t!]
  \caption{\textbf{General Vision-Language Understanding Results.} Best performance in \textbf{bold}.}
  \label{tab:general_ability}
  \vspace{-0.2cm}
  \centering
  {\fontsize{8pt}{11pt}\selectfont
   \resizebox{\linewidth}{!}{%
     \begin{NiceTabular}{l c c c c c c c c c c}[colortbl-like]
       \toprule[1pt]
       % \rowcolor{gray!20}
       \textbf{Models} &\textbf{\textsc{MME}} &\textbf{\textsc{POPE}} & \textbf{\textsc{SEED-Bench}} & \textbf{\textsc{AI2D}} & \textbf{\textsc{SQA-test}} & \textbf{\textsc{MMMUv}} & \textbf{\textsc{MMStar}} & \textsc{\textbf{HallusionBench}} \\
        \midrule
       SpatialRGPT-8B \cite{cheng2024spatialrgpt} & \textbf{1667}/348  & 85.50 & 67.00  & 67.42   & 81.81 & 41.40  & 43.98 & 40.80 &  \\
       \rowcolor{ourVLM}\textbf{\modelname{} \dponame{}} 8B   &     \textbf{1667}/\textbf{503}      &   \textbf{89.71}    &   \textbf{76.21}    &     \textbf{78.85}    &     \textbf{93.85}  &    \textbf{48.11}   & \textbf{55.43} & \textbf{51.10}    \\
       \bottomrule
     \end{NiceTabular}
   }
  }
\end{table}

%% file: assets/figures/qualfig.tex
\begin{figure}[!t]
    \centering    
    \includegraphics[width=0.99\linewidth]{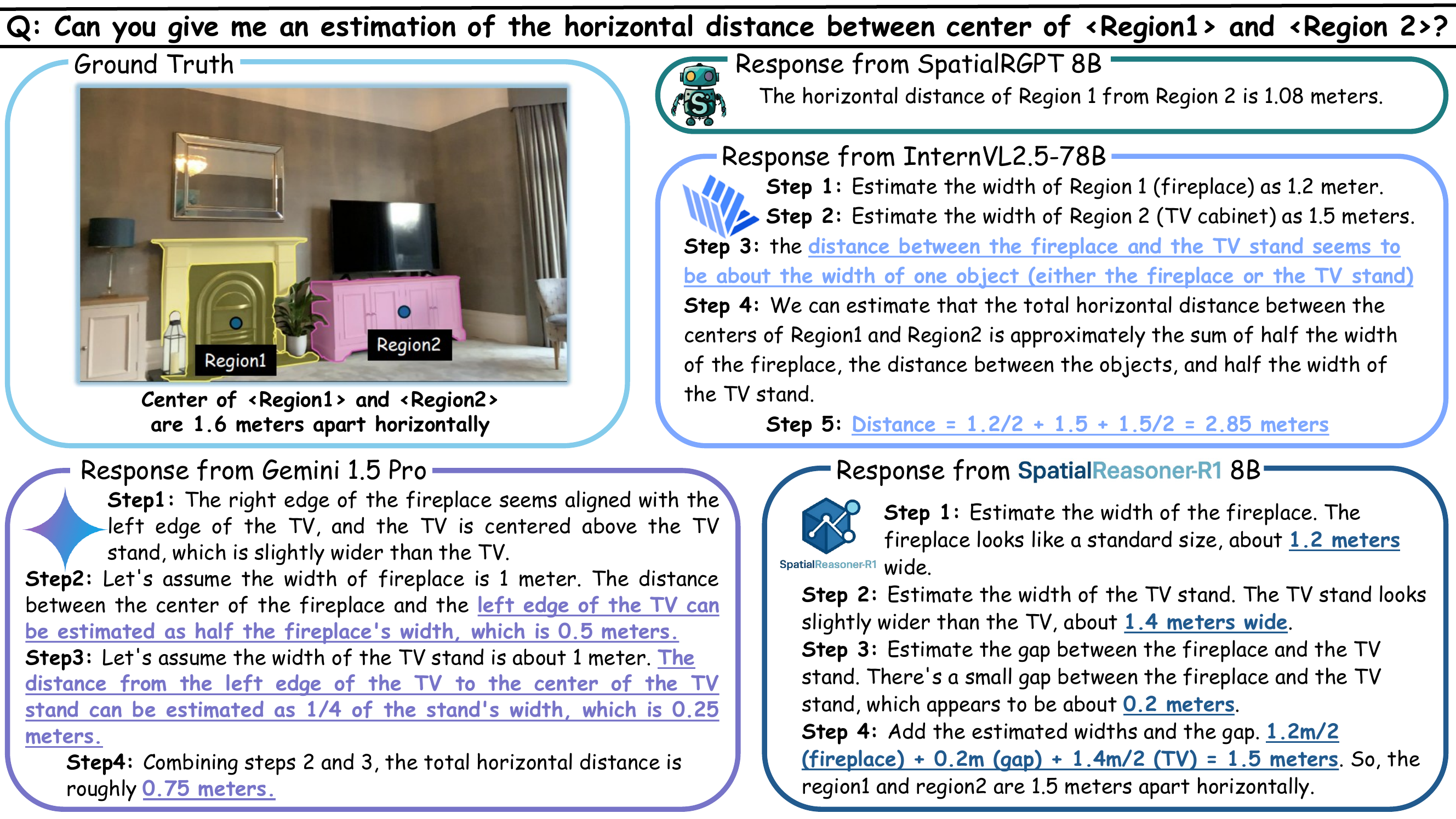}
    \vspace{-0.1cm}
    \caption{\textbf{Qualitative Examples of Spatial Reasoning Across Models.} \modelnamenc{} demonstrates a coherent, multi-step logical chain that closely matches the ground truth, while other models exhibit less precise or less interpretable reasoning paths.}
    \label{fig:qual}
\end{figure}

%% file: assets/tables/alpha_lambda_compact.tex
\begin{table}[t!]
  \centering 
  \begin{minipage}{0.48\textwidth}
      \centering 
      \captionsetup{justification=centering, singlelinecheck=false} 
      \caption{\textbf{Effect of Alpha (\(\alpha\)).}}
      \vspace{-0.1cm}
      \label{tab:alpha_compact}
      {\fontsize{8pt}{11pt}\selectfont
       \begin{NiceTabular}{l cccc}[colortbl-like]
         \toprule[1.2pt]
         \textbf{Metric}  & \textbf{10\%} & \textbf{20\%} & \textbf{30\%} & \textbf{40\%} \\
         \midrule[1.2pt]
         Direct Dist.     & 53.38 & 56.76 & \textbf{60.13} & 58.11 \\
         Horiz. Dist.     & 52.46 & 55.74 & \textbf{59.01} & 56.55 \\
         Vert. Dist.      & 65.09 & 67.92 & \textbf{71.75} & 69.81 \\
         Width            & 51.88 & 57.89 & \textbf{65.41} & 63.16 \\
         Height           & 56.39 & 57.14 & 57.89 & \textbf{58.64} \\
         Direction        & 91.58 & 92.23 & 92.52 & \textbf{94.39} \\
         \bottomrule[1.2pt]
       \end{NiceTabular}%
      }
  \end{minipage}
  \hfill
  \begin{minipage}{0.44\textwidth}
       \centering 
       \captionsetup{justification=centering, singlelinecheck=false} 
       \caption{\textbf{Effect of Lambda (\(\lambda\)) at \(\alpha\)=30\%.}}
       \vspace{-0.1cm}
      \label{tab:lambda_compact}
      {\fontsize{8pt}{11pt}\selectfont
       \begin{NiceTabular}{l cccc}[colortbl-like]
         \toprule[1.2pt]
         \textbf{Metric} & \textbf{\(\lambda\)=0.2} & \textbf{\(\lambda\)=0.4} & \textbf{\(\lambda\)=0.6} & \textbf{\(\lambda\)=0.8} \\
         \midrule[1.2pt]
         Direct Dist.    & 54.05 & 57.38 & \textbf{60.13} & 59.45 \\
         Horiz. Dist.    & 53.27 & 57.37 & \textbf{59.01} & 58.19 \\
         Vert. Dist.     & 65.09 & 68.86 & \textbf{71.75} & 69.81 \\
         Width           & 53.38 & 60.15 & \textbf{65.41} & 54.67 \\
         Height          & 56.39 & 57.14 & \textbf{57.89} & \textbf{57.89} \\
         Direction       & 91.58 & 91.58 & 92.53 & \textbf{93.45} \\
         \bottomrule[1.2pt]
       \end{NiceTabular}%
      }
  \end{minipage}
\end{table}

%% file: sections/05_conclusion.tex
\section{Conclusion}
\label{sec:conclusion}
In this work, we introduce \textbf{\modelname{}}, a novel VLM with state-of-the-art spatial reasoning capabilities, trained 
with a proposed fine-grained DPO (\dponame{}) method that decomposes LongCoT paths into description and reasoning components, allowing for targeted preference-based learning and enhanced logical reasoning. \dponame{} is guided by a set of comprehensive rewards that evaluate reasoning paths across visual consistency, spatial alignment, logical coherence, and depth-based verification.
Additionally, we propose a Multi-Model Monte Carlo Tree Search (\mctsname{}) strategy that leverages multiple VLMs to generate high-quality, diverse LongCoT data.
Our comprehensive evaluations demonstrate \modelnamenc{} achieves state-of-the-art performance, outperforming significantly larger models.
Moving forward, we plan to evaluate \dponame{} on additional VLM tasks such as GUI navigation and reasoning segmentation.  

\section*{Acknowledgments}
This research was supported by a gift from Google AI, the Google TPU Research Cloud (TRC) program, and the U.S. Defense Advanced Research Projects Agency (DARPA) under award numbers HR00112390062 and HR001125C0303. We gratefully acknowledge the cloud TPU credits from the Google TPU Research Cloud (TRC) program and the Google Tunix (Tune-in-JAX) team for their feedback.
The views and conclusions contained herein are those of the authors and should not be interpreted as necessarily representing the official policies, either expressed or implied, of Google, DARPA, or the U.S. Government. The U.S. Government is authorized to reproduce and distribute reprints for governmental purposes notwithstanding any copyright annotation therein.

%% file: sections/06_appendix.tex
\appendix
%\tableofcontents
%\clearpage
\section{Fine-Grained Spatial Reward Details}\label{appendix:rewards}
This appendix details design and hyperparameter choices for the fine-grained spatial rewards introduced in Section~\ref{sec:rlaif}. The prompt template for estimating rewards is shown in Figure~\ref{fig:llm_judge_prompt_part3}.

\subsection{Visual Consistency Reward ($\mathcal{R}_{\mathrm{vc}}$)}
The Visual Consistency Reward quantifies alignment between the generated description and the visual scene across four continuous criteria: Existence, Attribute accuracy, Completeness, and Appropriateness. Each component yields a score in the range $[0.0, 1.0]$, where its continuous range enables fine-grained assessment, permitting fractional scores when descriptions partially satisfy evaluation criteria. Scores near 0.0 indicate misalignment, while scores near 1.0 denote perfect alignment. Intermediate values reflect varying degrees of partial correctness or uncertainty. The total reward, $\mathcal{R}_{\mathrm{vc}} \in [0, 4.0]$, distinguishes varying degrees of alignment across responses.

\subsection{Depth-Guided Spatial Reward ($\mathcal{R}_{\mathrm{sp}}$)}
We introduce a depth-guided reward to evaluate the spatial accuracy of model outputs using ground-truth depth maps. The reward is computed independently for the {description}  $\bm{R}_{\text{desc}}$ and {reasoning} $\bm{R}_{\text{reason}}$ components, yielding two sub-scores: \(\mathcal{R}_{\text{sp,desc}}\) and \(\mathcal{R}_{\text{sp,reason}}\), each ranging from 0 to 4. These scores capture the alignment of spatial expressions with geometric cues in the image. The final spatial reward is given by $\mathcal{R}_{\mathrm{sp}}\!=\!\mathcal{R}_{\text{sp,desc}}+ \mathcal{R}_{\text{sp,reason}}$.

\textbf{Uncertainty Weight ($W_u$).} Spatial expressions in model outputs often include uncertain qualifiers. $W_u$ ranges from 0.8 to 1.0, with 1.0 indicating complete certainty in spatial assertions, and the lower bound of 0.8 representing cautious but plausible uncertainty. Setting the lower bound at 0.8 balances cautious language (\eg ``approximately,'' ``possibly'') without overly penalizing reasonable uncertainty. Lower values (below 0.8) would overly penalize reasonable, conservative predictions and discourage the model from producing cautious but informative reasoning.

\textbf{Context-aware Weight ($W_c$).} The context-aware weight $W_c \in [0.8, 1.0]$ reflects the relevance of spatial statements to the question. Explicitly asked spatial relationships are assigned $W_c\!=\!1.0$, while auxiliary or indirect spatial references are assigned $W_c\!=\!0.8$. This distinction prioritizes primary spatial relations explicitly required by the query, ensuring the model emphasizes essential spatial assertions more significantly. Scores below 0.8 would disproportionately underemphasize auxiliary information, degrading the model's ability to handle broader contextually relevant details.

Given a response, we extract all spatial relationship statements from the description and reasoning response sections. Each statement is then evaluated using GPT-4o by comparing the original image and its corresponding depth image, which is generated using Depth Anything~\cite{yang2024depth}, to obtain a correctness score $r_i \in [0, 1]$. Every statement is also assigned an associated $W_u^{(i)}$ and $W_c^{(i)}$. The spatial reward scores are computed as
\begin{equation}
\mathcal{R}_{\text{sp,desc}} = \frac{1}{n} \sum_{i=1}^{n} W_u^{(i)} \cdot W_c^{(i)} \cdot r_i,
\qquad
\mathcal{R}_{\text{sp,reason}} = \frac{1}{m} \sum_{i=1}^{m} W_u^{(i)} \cdot W_c^{(i)} \cdot r_i,
\end{equation}
where $n$ and $m$ denote the number of spatial statements in the description and reasoning components, respectively, and $r_i$ represents the correctness of each spatial relationship, validated against both the RGB image and its corresponding depth map.

\subsection{Logical Coherence Reward ($\mathcal{R}_{\mathrm{lc}}$)}
This reward quantifies the logical robustness of a response by aggregating four components: Factual Consistency, Logical Coherence, Correct Rule Application, and Conclusion Validity. Each component is scored in the range $[0.0, 1.0]$, with fractional values capturing partial correctness, \eg from minor gaps in logical sequences to partial inaccuracies in applying physical, spatial, or logical rules. Scores of 0.0 and 1.0 indicate complete logical coherence failure or perfect logical chains, respectively. The final reward, $\mathcal{R}{\mathrm{lc}} \in [0, 4.0]$, reflects the overall logical quality of the reasoning.

\section{Structured Output Format Specification for \mctsname{}} 
\label{appendix:MMCTS_format} 
To enable reliable parsing and downstream analysis, \mctsname{} requires reasoning paths from VLMs to follow a standardized structured format. This format uses Markdown-style headings to clearly segment key components of the reasoning trace. Each section begins with a line prefixed by \texttt{\#\#\#}, followed by a descriptive heading. 
The defined sections are:
\begin{itemize}[itemsep=0.5ex, parsep=0pt, topsep=0pt, leftmargin=0.1cm]
    \item[] \texttt{\#\#\# Description}: Details the input context, such as the visual scene or scenario description.
    \item[] \texttt{\#\#\# Rationale}: Summarizes the overall reasoning strategy or justification. 
    \item[] \texttt{\#\#\# Let's think step by step}: An optional phrase before detailed step-by-step breakdown. 
    \item[] \texttt{\#\#\# Step N} (\eg \texttt{\#\#\# Step 1}, \texttt{\#\#\# Step 2}, ...): Enumerates the sequential steps involved in the reasoning procedure. Multiple steps are typically present.
    \item[] \texttt{\#\#\# In Conclusion}: States the final derived conclusion of the reasoning process.
\end{itemize}
Figure~\ref{fig:m3cts} shows the reasoning tree example produced by \mctsname{}.

\section{Node Evaluation Protocol for \mctsname{}}
\label{appendix:MMCTS_evaluation}

To ensure the semantic and visual quality of each candidate reasoning step \( s_{k,t} \) within the \mctsname{} framework, we employ a structured multi-criteria evaluation. Individual steps that form a reasoning path are segmented by \textbf{\texttt{\#\#\#}}. Each candidate \( s_{k,t} \in \mathcal{S}_c \) is independently evaluated by two multimodal models, Gemini 1.5 Pro and Qwen2.5VL-72B, along key distinct dimensions:
\begin{enumerate}[itemsep=0.5ex, parsep=0pt, topsep=0pt, leftmargin=0.5cm]
    \item[{\Large$\diamond$}] \textbf{Visual Description Accuracy}: 
    Assesses whether the entities, attributes, and contextual cues described in \( s_{k,t}\) correctly reflect the visual content of the input image. This includes references to objects, colors, spatial layouts, and contextual cues.
    \item[{\Large$\diamond$}] \textbf{Spatial Consistency}: Evaluates whether the spatial relations expressed in \( s_{k,t} \) (\eg ``above,'' ``to the left of,'' ``behind'') are consistent with both the RGB image and depth map generated via the Depth Anything model~\cite{yang2024depth}. Errors such as inversion of relations (\eg stating ``behind'' instead of ``in front'') are penalized.  
    \item[{\Large$\diamond$}] \textbf{Logical Reasoning Coherence}: For steps within the ``think step-by-step'' chain-of-thought reasoning block, this component checks whether the logical flow of inferences is coherent and justified. This includes identifying unsupported jumps in logic or contradictions.
\end{enumerate}

Each criterion  \(\mathbb{I}^{(m)}_{\text{eval}}(s_{k,t})\) is rated as follows:
\[
 \mathbb{I}^{(m)}_{\text{eval}}(s_{k,t}) =
\begin{cases}
+1, & \text{if the content is entirely accurate according to model } m; \\\\
\;\;\;0, & \text{if the content is ambiguous or partially accurate;} \\\\
-1, & \text{if there is any clear inaccuracy.}
\end{cases}
\]

We preserve high-quality paths by pruning the candidate set. Specifically,  we retain any node \( s_{k,t} \) whose aggregated score across all evaluators and criteria is non-negative, \ie
$\mathcal{S}_{\text{c}}^*\!=\!\{ s_{k,t} \mid R(s_{k,t}) \geq 0 \}.$
This threshold is chosen empirically to balance filtering out incorrect steps while maintaining adequate reasoning diversity.

\section{Training Details}
\label{appendix:training}

\subsection{Implementation Details}
\modelnamenc{} is built upon the Sa2VA architecture \cite{yuan2025sa2va}, which is based on InternVL2.5. We train the 8B-parameter model in two stages on two NVIDIA H100 GPUs, each stage taking approximately 2.5 days. For supervised fine-tuning, we employ AdamW optimizer with a learning rate of $4\times10^{-5}$, weight decay of 0.05, and a 5\% linear warm-up schedule, using a batch size of 2 per device with gradient accumulation over 4 steps. For Direct Preference Optimization, we similarly use AdamW with learning rate of $1\times10^{-7}$, weight decay of 0.05, and a 5\% warm-up, training with a batch size of 1 per device.

\input{assets/figures/m3cts}

\subsection{Training Data} \label{appendix:dataset}

For SFT, we convert samples from the \textsc{Open Spatial} dataset~\cite{cheng2024spatialrgpt} to reasoning chains using the \mctsname{} pipeline. While the original \textsc{Open Spatial} dataset provides single-sentence answers, we transform 400K samples, grounded in distinct images, into structured  LongCoT reasoning chains, where examples are used to teach the model to generate high-quality, step-by-step spatial reasoning responses.
For Direct Preference Optimization (DPO) training, the goal is to train the model to distinguish high-quality spatial reasoning from suboptimal or subtly flawed alternatives. To this end, we utilize our \textsc{Open Spatial Reasoning} dataset, described below, that consists of spatial reasoning preference pairs. 

An additional set of 100K  challenging negative pairs is meticulously crafted by perturbing only the conclusion keywords of high-quality positive samples. Each original response represents a coherent and accurate reasoning path with a factually correct outcome. To create the corresponding negative sample, we retain the exact description and reasoning segments and alter only the final conclusion value. This yields tightly controlled preference pairs that isolate correctness at the conclusion level. For example, a positive sample may assert \texttt{``The distance between region1 and region2 is 11 meters.''}, while its negative perturbed counterpart is \texttt{``The distance between region1 and region2 is 10 meters.''}

Our method adopts a data-centric strategy that emphasizes high-quality supervision and reasoning diversity. Instead of collecting large volumes of weakly aligned or noisy data, we curate training examples using the \mctsname{} sampling strategy guided by structured reward evaluations. By applying reward-based filtering, we reduce noise and enforce a consistent output structure. In parallel, using multiple VLMs during generation introduces variation in reasoning styles, improving coverage of diverse spatial patterns and edge cases. The effectiveness of this approach is evident in the substantial performance gains of DPO-trained models over their simpler SFT counterparts (Table~\ref{tab:spatial_ability}) and the reasoning improvements and diversity depicted in Figure \ref{fig:m3cts}.

\subsection{Open Spatial Reasoning Dataset}
We curate the \textsc{Open Spatial Reasoning} dataset, a collection of 400K Vision Question Answering (VQA) preference pairs $(y_p, y_l)$, to support training of preference-based spatial reasoning models. This dataset is derived from the \textsc{Open Spatial} dataset~\cite{cheng2024spatialrgpt}, which provides image-based spatial questions paired with ground-truth answers and offers 10 question variations per image-grounding scenario.
To construct each preference pair, we randomly sample a question instance from the source dataset, and generate a diverse pool of eight candidate answers using four distinct sources: our \mctsname{} pipeline, Gemini 1.5 Pro, GPT-4o, and our \modelnamenc{} Supervised Fine-Tuned (SFT) model, with each method contributing two response variants. All eight candidate responses are independently evaluated by our fine-grained spatial reward mechanism (Appendix~\ref{appendix:rewards}). The highest-scoring response is selected as the preferred answer $(y_p)$, while the response with the lowest score is designated as the less-preferred $(y_l)$, ensuring that each preference pair is anchored in meaningful fine-grained spatial reasoning quality. Figure~\ref{fig:dataset} shows dataset examples. 
\input{assets/figures/dataset}

\section{Evaluation Details}
\label{appendix:benchmark}
We evaluate on \textsc{SpatialRGPT-Bench}~\cite{cheng2024spatialrgpt}, a benchmark specifically designed to assess the 3D spatial reasoning abilities of VLMs, featuring 657 qualitative and 749 quantitative VQA pairs, covering 88 object classes across diverse environments.  We employ the same GPT-4 evaluation proposed in \textsc{SpatialRGPT-Bench}~\cite{cheng2024spatialrgpt} for evaluating the free-form responses generated by the models. For \textit{qualitative} questions, GPT-4o assesses the semantic alignment between the model's response and the ground-truth answer, assigning a binary score (1 for correct, 0 for incorrect). For \textit{quantitative} questions (\eg distance, size), GPT-4o first extracts numerical values from both the prediction and the ground truth, standardizing them to a common unit (meters). We then compute accuracy (\eg success rate defined as predictions within $\pm 25\%$ of the ground truth).

We also evaluate on several general vision-language benchmarks to provide a comprehensive assessment of \modelnamenc{}'s capabilities. Specifically, we use \textsc{MME}~\cite{li2023evaluating} to assess multimodal models on perception and cognition tasks across a wide range of domains. \textsc{POPE}~\cite{li2023evaluating} is employed to evaluate object hallucination in testing the ability of VLMs to ground responses to visual content, while \textsc{SEED-Bench}~\cite{li2023seed} offers a multi-dimensional evaluation, covering aspects from image understanding to complex reasoning across various modalities and tasks. We further utilize \textsc{AI2D}~\cite{kembhavi2016diagram}, a benchmark focusing on diagram understanding and reasoning, which requires parsing visual elements and their relationships within schematic representations. \textsc{SQA}~\cite{lu2022learn} is used to measure the model's ability to answer science-related questions based on visual context, often requiring domain-specific knowledge and reasoning. \textsc{MMMU}~\cite{yue2024mmmu} evaluates massive multi-disciplinary multimodal understanding and reasoning across diverse college-level subjects. Moreover, \textsc{MMStar}~\cite{chen2024we} provides a challenging benchmark with meticulously curated, multimodal instances that require advanced reasoning, low hallucination, and resistance to leading questions. Finally, \textsc{HallusionBench}~\cite{guan2024hallusionbench} is specifically designed to quantitatively measure and analyze the hallucination phenomena in VLMs, probing for both object-level and attribute-level inconsistencies.

\input{assets/figures/additionexample1}
\section{Qualitative Experiment Examples}\label{appendix:examples}
In this section, we provide additional qualitative experiment examples. Figure~\ref{fig:additionexample1} shows a question that requires estimation of the horizontal distance between a truck and a pedestrian. \textbf{\modelname{}} demonstrates a clear advantage by decomposing the scene into semantically meaningful components, explicitly reasoning over the widths of multiple traffic lanes, the roadside, and the sidewalk. This results in an estimated distance that closely matches the ground truth and provides full transparency into the model’s stepwise deductions. In contrast, \textbf{InternVL2.5-78B} bases its answer primarily on the width of the trucks and the space between them, omitting the crucial step of accounting for the distance from the pedestrian to the roadway, which leads to significant underestimation. \textbf{Gemini1.5Pro} correctly recognizes that the separation includes the truck, traffic lane, and sidewalk, but substantially underestimates the width of the sidewalk, causing a notable error in its final answer. Meanwhile, \textbf{SpatialRGPT-8B} provides a more accurate estimate than Gemini or InternVL2.5-78B, but still has a gap compared to the ground truth. Most importantly, it cannot generate step-by-step reasoning traces.\looseness-1

Figure~\ref{fig:additionexample2} presents another illustrative example evaluating spatial reasoning capabilities of various models, specifically focusing on size comparison between two highlighted image regions. The question is whether Region 1 (a computer monitor) appears smaller than Region 2 (a computer tower). \textbf{\modelname{}} accurately identifies Region 2 as a computer tower and explicitly reasons by comparing Region 1 with the closest computer tower positioned adjacent to the monitor. This systematic visual grounding and clear comparative reasoning enable \modelnamenc{} to correctly conclude that the monitor is indeed smaller than the tower. By contrast, the baseline models exhibit varying degrees of errors and reasoning inadequacies. \textbf{InternVL2.5-78B} relies significantly on prior general knowledge about typical object dimensions and incorrectly concludes the monitor is not smaller, without effectively validating this against the visual evidence provided. textbf{Gemini1.5Pro} fails entirely to recognize what object Region 2 represents, causing it to inaccurately rely purely on the objects' visual proximity and perspective, leading to an incorrect conclusion. Lastly, the \textbf{SpatialRGPT-8B} model directly presents an incorrect judgment (``Region1 is not smaller'') without providing any interpretable reasoning steps or visual grounding.
\input{assets/figures/additionexample2}

We also provide a failure example in this section. Figure~\ref{fig:fail_case} illustrates a representative failure case on vertical size estimation in an indoor setting. The query is “How tall is Region 1?”. Here, Region 1 corresponds to the dresser mirror on the right, adjacent to a sleigh‑style bed headboard on the left. \modelname{} produces an estimate of approximately 2.0m, while the ground truth is closer to 1.5m.
Our model’s reasoning proceeds as follows: (1) segments Region 1 (the mirror) and searches for a nearby object of familiar scale,  (2) identifies the bed headboard and assumes a typical headboard height of 1.5m, further estimating that the mirror extends about 0.5m above the headboard, and (3) sums these values to obtain about 2.0m.
This error arises from overreliance on default furniture priors rather than fully grounding the estimate in image evidence, such as the mirror’s vertical extent relative to the floor plane and its contact points with the dresser. To mitigate this, training incorporates fine‑grained reward signals that explicitly reward consistency between predicted measurements and image‑derived cues, encouraging verification of intermediate steps (e.g., floor contact, vanishing‑line alignment) before finalizing a measurement.
\input{assets/figures/fail_case}

\section{Broader Impacts}
This work aims to improve the spatial reasoning capabilities of vision-language models through fine-grained preference optimization. Accurate spatial understanding is critical for downstream applications such as robotics, autonomous navigation, assistive technologies, and visual analytics. By introducing more interpretable and structured reasoning mechanisms, our method can contribute to building AI systems that are safer, more transparent, and more aligned with human expectations in spatially grounded tasks.
However, as with other vision-language systems, potential risks remain. If deployed in safety-critical domains, incorrect spatial inferences, especially in edge cases, could lead to unintended consequences. Additionally, reward scoring and generation rely on foundation models that may encode hidden biases, which can propagate through the training pipeline. Although we attempt to mitigate these risks via multi-source sampling and structured evaluation, future work should explore robustness to distribution shifts, adversarial spatial prompts, and the inclusion of human-in-the-loop verification for high-stakes use cases.

\section{Limitations} 
While our work demonstrates strong improvements in spatial reasoning, a limitation of our approach is its reliance on explicit region representations provided as input to disambiguate object references within the spatial queries. Enabling the model to implicitly ground entities solely based on natural language descriptions remains an avenue for future investigation, which would enhance the model's flexibility in real-world scenarios. Future work could focus on integrating implicit linguistic context understanding to alleviate this constraint. Finally, our focus is limited to 2D spatial reasoning; extending this framework to 3D or embodied contexts would require structural adjustments left for future work.\looseness-1

\newpage
\thispagestyle{empty}

\begin{figure}[t!] 
\vspace{-0.9cm}
\centering
\resizebox{0.95\textwidth}{!}{
\begin{planbox}{System Prompt for LongCoT Reward Evaluation} 
\small

The following is a spatial reasoning task, and this is the question: \texttt{{question}} and the ground truth is: \texttt{{ground\_truth}}. The response is divided into different sections.
There are 4 dimensions to evaluate, and I will provide you with the corresponding image and text for reference. You will need to evaluate the response based on the following criteria: \\

\textbf{The first task: Descriptive Scoring (Total 0–4.0 points)} \\
   Evaluate the "Description" section based on:
      \begin{itemize} \itemsep0em
          \item \textbf{Existence:} Assign scores from 0 to 1.0, where 1.0 means mostly confidently correct and 0 means mostly confidently incorrect. Does the description correctly identify objects that actually appear in the image?
          \item \textbf{Attribute Accuracy:} Assign scores from 0 to 1.0, where 1.0 means mostly confidently correct and 0 means mostly confidently incorrect. Are the object's attributes (color, shape, size, etc.) described accurately?
          \item \textbf{Completeness:} Assign scores from 0 to 1.0, where 1.0 means mostly confidently correct and 0 means mostly confidently incorrect. Does the description include all key objects and necessary details relevant to the question?
          \item \textbf{Appropriateness:} Assign scores from 0 to 1.0, where 1.0 means mostly confidently correct and 0 means mostly confidently incorrect. Does the description focus on the core aspects of the question?
      \end{itemize}
   Clearly state the score for each sub-category and sum them to obtain the final descriptive score. You need to give the score with the following format: 
   \texttt{{\{"task1\_score": your score\}}}\\

\textbf{The second task: Depth-Guided Spatial Relationship Scoring – Description (Total 0–4.0 points)} \\
   Evaluate all spatial statements within the "Description" section using the provided depth image as ground truth.
      \begin{itemize} \itemsep0em
            \item For each spatial claim in the description:
                \begin{itemize} \itemsep0em
                    \item \textbf{Correctness score:} Assign 1 if the spatial claim is correct based on the depth image, and 0 if not correct.
                    \item \textbf{Uncertainty score:} For claims expressed with uncertainty (using words like "approximately", "roughly", "possibly"), assign a score from 0.8 to 1.0, where 1.0 means the statement is expressed with high certainty.
                    \item \textbf{Relationship score:} Assign a weight from 0.8 to 1.0 based on whether the relationship is explicitly emphasized by the question (1.0) or is extra/irrelevant information (0.8).
                \end{itemize}
      \end{itemize}
   Provide a detailed breakdown for each spatial claim. Calculate the final score as: (Sum of (Correctness score × Uncertainty score × Relationship score)) / (Number of claims), then scale to 4.0. \\
   You need to give the score with the following format: \\
   \texttt{{\{"task2\_claim\_score": [Correctness score, Uncertainty score, Relationship score]\}}} \\

\textbf{The third task: Depth-Guided Spatial Relationship Scoring – Reasoning (Total 0–4.0 points)} \\
   Apply the same evaluation method as in Task 2 to the spatial statements within the "Reasoning" section.
      \begin{itemize} \itemsep0em
           \item For each spatial claim in the reasoning:
                \begin{itemize} \itemsep0em
                    \item \textbf{Correctness score:} Assign 1 if the spatial claim is correct based on the depth image, and 0 if not correct.
                    \item \textbf{Uncertainty score:} For claims expressed with uncertainty, assign a score from 0.8 to 1.0, where 1.0 means the statement is expressed with high certainty.
                    \item \textbf{Relationship score:} Assign a weight from 0.8 to 1.0 based on whether the relationship is explicitly emphasized by the question (1.0) or is extra/irrelevant information (0.8).
                 \end{itemize}
      \end{itemize}
   Provide a detailed breakdown for each spatial claim. Calculate the final score as: (Sum of (Correctness score × Uncertainty score × Relationship score)) / (Number of claims), then scale to 4.0. \\
   You need to give the score with the following format: \\
   \texttt{{\{"task3\_claim\_score": [Correctness score, Uncertainty score, Relationship score]\}}}\\

\textbf{The fourth task: Reasoning Scoring (Total 0–4.0 points)} \\
   Evaluate the "Reasoning" section (the chain-of-thought) based on:
      \begin{itemize} \itemsep0em
          \item \textbf{Factual Consistency:} Assign scores from 0 to 1.0, where 1.0 means mostly confidently correct and 0 means mostly confidently incorrect. Are the claims consistent with the image, depth image, and the earlier description?
          \item \textbf{Logical Coherence:} Assign scores from 0 to 1.0, where 1.0 means mostly confidently correct and 0 means mostly confidently incorrect. Do the reasoning steps flow logically without gaps or contradictions?
          \item \textbf{Correct Application of Rules:} Assign scores from 0 to 1.0, where 1.0 means mostly confidently correct and 0 means mostly confidently incorrect. Are physical, spatial, and logical rules applied correctly?
          \item \textbf{Conclusion Validity:} Assign scores from 0 to 1.0, where 1.0 means mostly confidently correct and 0 means mostly confidently incorrect. Does the reasoning properly support the final answer?
      \end{itemize}
   Clearly state the score for each sub-category and sum them to obtain the final reasoning score. \\
   You need to give the score with the following format: 
   \texttt{{\{"task4\_score": your score\}}}

\end{planbox}
}
\vspace{-0.2cm}
\caption{\textbf{System Prompt for Evaluating LongCoT Spatial Reasoning} w.r.t. descriptive accuracy, spatial alignment, and logical consistency of reasoning steps.}
\label{fig:llm_judge_prompt_part3}
\end{figure}

%% file: assets/figures/m3cts.tex
\begin{figure}[!t]
    \centering    \includegraphics[width=\linewidth]{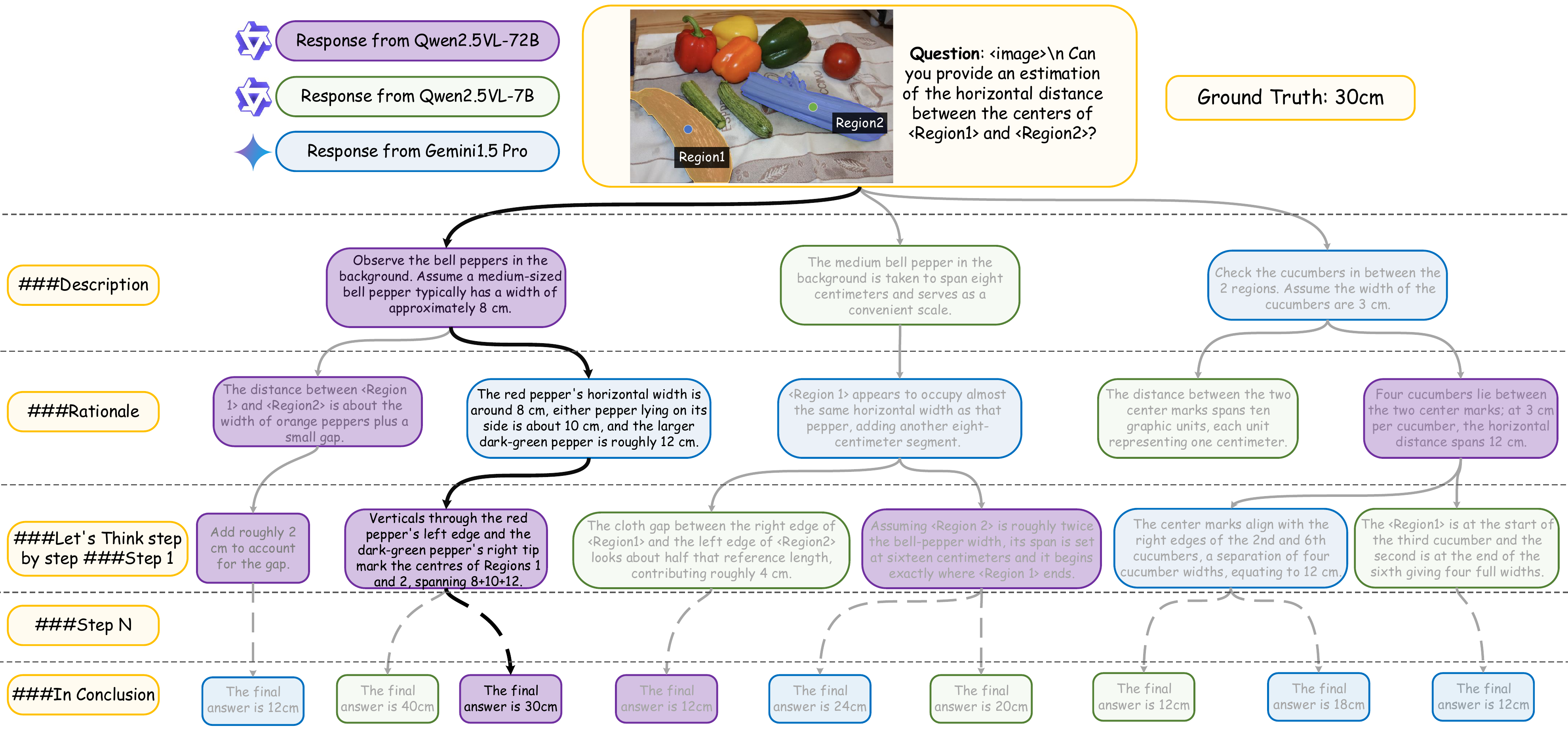}
    \caption{\textbf{Example Reasoning Tree from the M3CTS Data Generation Pipeline.} Diverse candidate reasoning paths are sampled from multiple models. Each path follows a structured LongCoT format with markdown-style section headers that decompose the answer into interpretable reasoning stages.}
    \label{fig:m3cts}
\end{figure}

%% file: assets/figures/dataset.tex
\begin{figure}[!t]
    \centering    \includegraphics[width=0.99\linewidth]{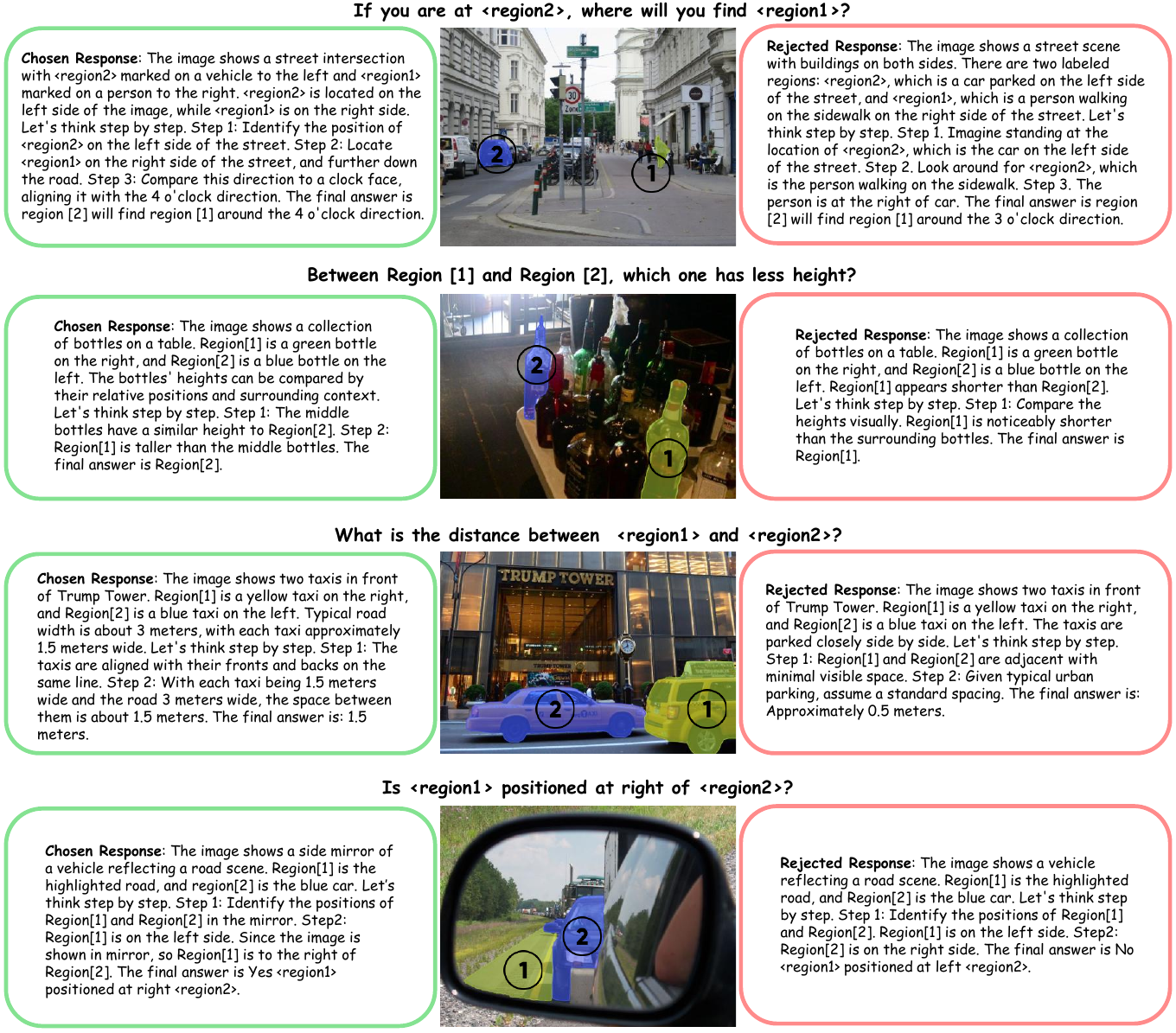}
    \vspace{-0.1cm}
    \caption{\textbf{Example DPO Pairs of our \textsc{Open Spatial Reasoning} Dataset,} constructed from \mctsname{}-generated reasoning trajectories. Each pair consists of a preferred and a rejected response to the same spatial question. The examples highlight differences in descriptive accuracy, spatial alignment, and reasoning coherence, which guide preference optimization during training.}
    \label{fig:dataset}
\end{figure}

%% file: assets/figures/additionexample1.tex
\begin{figure}[t]
    \centering    
    \includegraphics[width=0.99\linewidth]{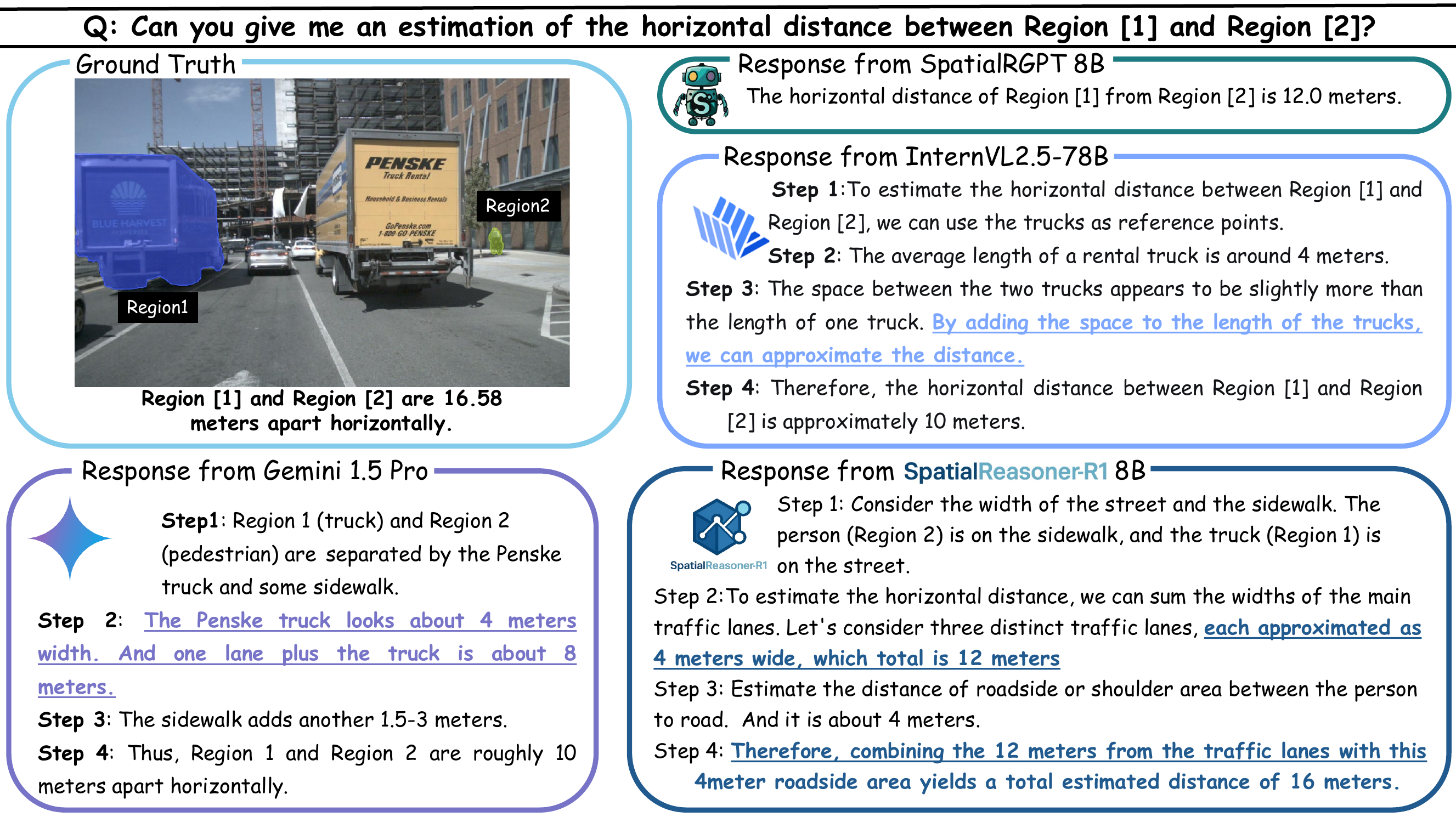}
    \vspace{-0.1cm}
    \caption{\textbf{Qualitative Examples of Spatial Reasoning Across Models.} \modelname{} demonstrates coherent, step-by-step spatial reasoning that closely aligns with ground truth estimates. In contrast, baseline models produce less precise or partially incorrect reasoning steps, often neglecting key visual cues or misestimating spatial references.}
    \label{fig:additionexample1}
    \vspace{-0.1cm}
\end{figure}

%% file: assets/figures/additionexample2.tex
\begin{figure}[t!]
    \centering    
    \includegraphics[width=0.99\linewidth]{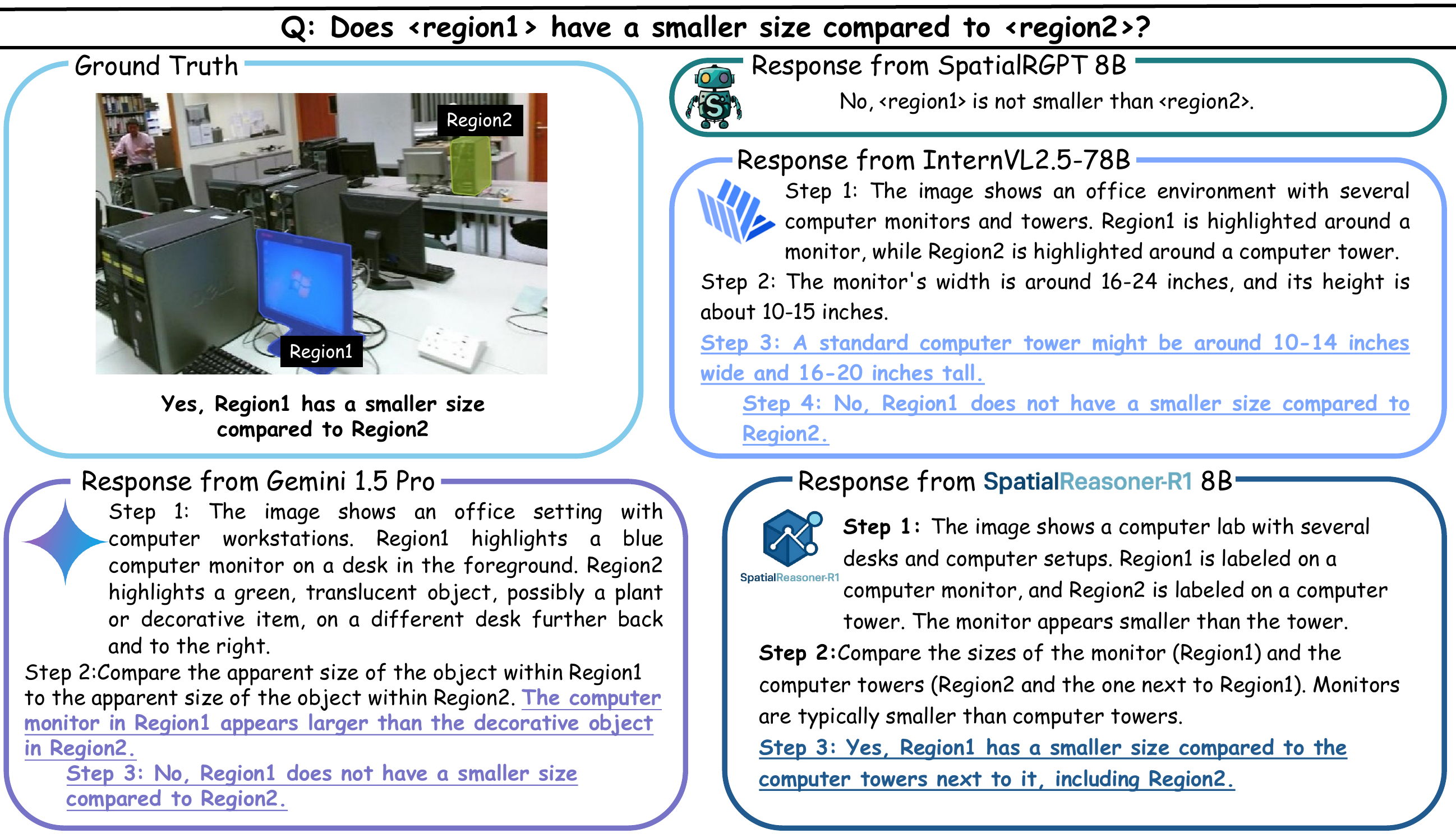}
    \vspace{-0.1cm}
    \caption{\textbf{Qualitative Examples of Spatial Reasoning Across Models.} \modelname{} correctly recognizes Region2 as a computer tower and compares it clearly with the nearby monitor, reaching an accurate conclusion. InternVL2.5-78B relies on general object size knowledge but provides incorrect reasoning, Gemini1.5Pro fails to identify Region2 clearly and draws incorrect visual conclusions, while SpatialRGPT-8B directly provides a wrong answer.}
    \label{fig:additionexample2}
\end{figure}

%% file: assets/figures/fail_case.tex
\begin{figure}[t!]
    \centering    
    \includegraphics[width=\linewidth]{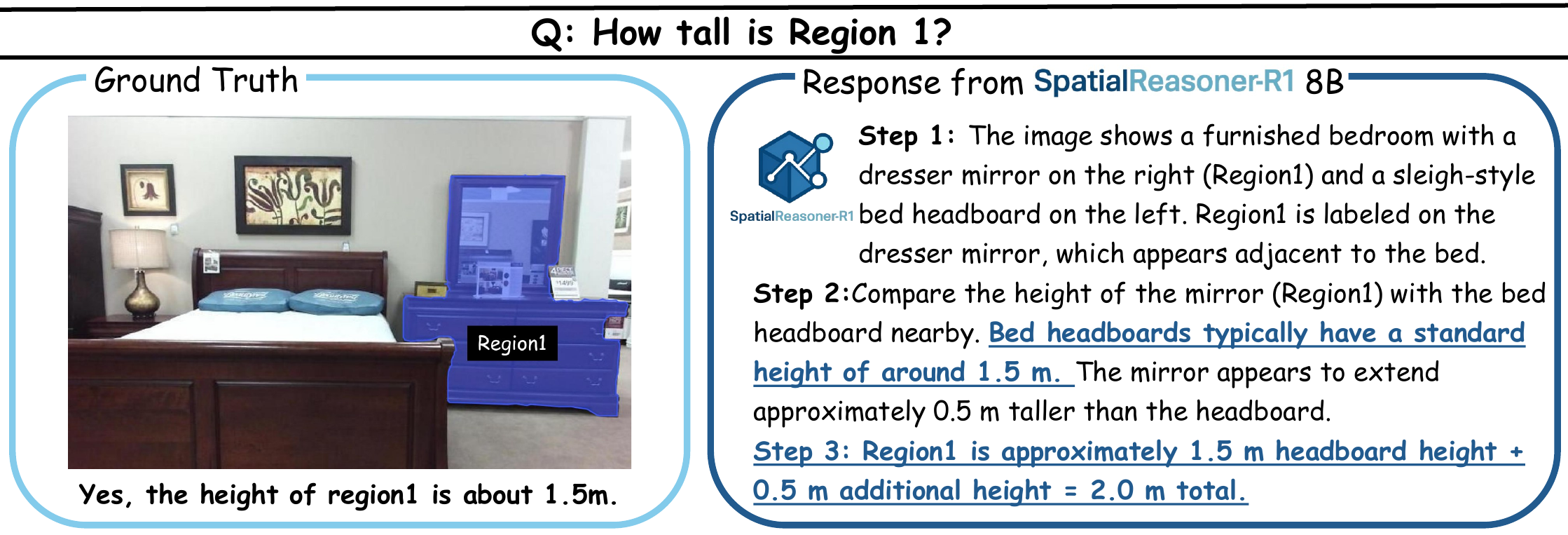}
    \caption{\textbf{Failure case} on height estimation in a furnished bedroom.}
    \label{fig:fail_case}
\end{figure}